\documentclass{article} 
\usepackage{nips12submit_e,times}

\pdfoutput=1

\nipsfinalcopy 

\author{
Alexandra Carpentier\\
Statistical Laboratory, CMS\\
Wilberforce Road, Cambridge\\
CB3 0WB UK \\
\texttt{a.carpentier@statslab.cam.ac.uk} \\
\And
R\'emi Munos \\
INRIA Lille - Nord Europe \\
40, avenue Halley\\
59000 Villeneuve d'ascq, France\\
\texttt{remi.munos@inria.fr} \\
}

\usepackage{epsf}
\usepackage{graphicx}
\usepackage{amsmath}
\usepackage{amssymb}
\usepackage{algorithmic}
\usepackage{algorithm}

\newcommand{\B}{\mathcal B}
\newcommand{\V}{\mathbb V}
\renewcommand{\P}{\mathbb P}

\newlength{\minipagewidth}
\newcommand{\bookbox}[1]{\small
\par\medskip\noindent
\framebox[\columnwidth]{
\begin{minipage}{\minipagewidth} {#1} \end{minipage} } \par\medskip }

\newcommand{\beq}{\begin{equation}}
\newcommand{\eeq}{\end{equation}}

\newcommand{\beqa}{\begin{eqnarray}}
\newcommand{\eeqa}{\end{eqnarray}}

\newcommand{\beqan}{\begin{eqnarray*}}
\newcommand{\eeqan}{\end{eqnarray*}}

\newcommand{\E}{\mathbb{E}}

\newcommand{\alg}{\mathcal A}

\newcommand{\hmu}{\hat{\mu}}

\newcommand{\var}{\sigma^2}

\newcommand{\si}{\sigma}
\newcommand{\hsi}{\hat{\sigma}}

\newcommand{\N}{\mathcal N}

\newcommand{\ind}[1]{\mathbb I\left\lbrace {#1} \right\rbrace}

\newcommand{\X}{\mathcal X}

\newcommand{\R}{\mathbb R}

\newtheorem{theorem}{Theorem}
\newtheorem{assumption}{Assumption}

\newtheorem{lemma}{Lemma}
\newtheorem{corollary}{Corollary}

\title{Adaptive Stratified Sampling for Monte-Carlo integration of Differentiable functions}

\begin{document}

\maketitle

\begin{abstract}
We consider the problem of adaptive stratified sampling for Monte Carlo integration of a differentiable function given a finite number of evaluations to the function. We construct a sampling scheme that samples more often in regions
 where the function oscillates more, while allocating the samples such that they are well spread on the domain (this notion shares similitude with low discrepancy). We prove that the estimate returned by the algorithm is almost similarly accurate as the estimate that an optimal oracle strategy (that would know the variations of the function \textit{everywhere}) would return, and provide a finite-sample analysis.
\end{abstract}

\vspace{-0.2cm}
\section{Introduction}
\vspace{-0.2cm}
In this paper we consider the problem of numerical integration of a differentiable function $f:[0,1]^d\rightarrow \R$ given a finite budget $n$ of evaluations to the function that can be allocated sequentially.

A usual technique for reducing the mean squared error (w.r.t.~the integral of $f$) of a Monte-Carlo estimate is the so-called stratified Monte Carlo sampling, which considers sampling into a set of strata, or regions of the domain, that form a partition, i.e.~a stratification, of the domain (see \cite{rubinstein2008simulation}[Subsection 5.5] or \cite{glasserman2004monte}). It is efficient (up to rounding issues) to stratify the domain, since when allocating to each stratum a number of samples proportional to its measure, the mean squared error of the resulting estimate is always smaller or equal to the one of the crude Monte-Carlo estimate (that samples uniformly the domain).


Since the considered functions are differentiable, if the domain is stratified in $K$ hyper-cubic strata of same measure and if one assigns uniformly at random $n/K$ samples per stratum, the mean squared error of the resulting stratified estimate is in $O(n^{-1}K^{-2/d})$. We deduce that if the stratification is built \textit{independently} of the samples (before collecting the samples), and if $n$ is known from the beginning (which is assumed here), the minimax-optimal choice for the stratification is to build $n$ strata of same measure and minimal diameter, and to assign only one sample per stratum uniformly at random. We refer to this sampling technique as Uniform stratified Monte-Carlo. The resulting estimate has a mean squared error of order $O(n^{-(1+2/d)})$. The arguments that advocate for stratifying in strata of same measure and minimal diameter are closely linked to the reasons why quasi Monte-Carlo methods, or low discrepancy sampling schemes are efficient techniques for integrating smooth functions. See~\cite{niederreiter1978quasi} for a survey on these techniques.

It is minimax-optimal to stratify the domain in $n$ strata and sample one point per stratum, but it would also be interesting to adapt the stratification of the space with respect to the function $f$. For example, if the function has larger variations in a region of the domain, we would like to discretize the domain in smaller strata in this region, so that more samples are assigned to this region. Since $f$ is initially unknown, it is not possible to design a good stratification before sampling. However an efficient algorithm should allocate the samples in order to estimate online the variations of the function in each region of the domain while, \textit{at the same time}, allocating more samples in regions where $f$ has larger local variations.
\\
The papers~\cite{A-EtoJou10, Grover, MC-UCB} provide algorithms for solving a similar trade-off when the stratification is fixed: these algorithms allocate more samples to strata in which the function has larger variations. It is, however, clear that the larger the number of strata, the more difficult it is to allocate the samples almost optimally in the strata. 

{\bf Contributions:} We propose a new algorithm, Lipschitz Monte-Carlo Upper Confidence Bound (LMC-UCB), for tackling this problem. It is a two-layered algorithm. It first stratifies the domain in $K \ll n$ strata, and then allocates uniformly to each stratum an initial small amount of samples in order to estimate roughly the variations of the function per stratum. Then our algorithm sub-stratifies each of the $K$ strata according to the estimated local variations, so that there are in total approximately $n$ sub-strata, and allocates one point per sub-stratum. In that way, our algorithm discretizes the domain into more refined strata in regions where the function has higher variations. It cumulates the advantages of quasi Monte-Carlo and adaptive strategies.

More precisely, our contributions are the following:

\begin{itemize}
  \item We prove an asymptotic lower bound on the mean squared error of the estimate returned by an optimal oracle strategy that has access to the variations of the function $f$ everywhere and would use the best stratification of the domain with hyper-cubes (possibly of heterogeneous sizes). This quantity, since this is a lower-bound on any oracle strategies, is smaller than the mean squared error of the estimate provided by Uniform stratified Monte-Carlo (which is the non-adaptive minimax-optimal strategy on the class of differentiable functions), and also smaller than crude Monte-Carlo.
  \item We introduce the algorithm LMC-UCB, that sub-stratifies the $K$ strata in hyper-cubic sub-strata, and samples one point per sub-stratum. The number of sub-strata  per stratum is linked to the variations of the function in the stratum. 
 We prove that algorithm LMC-UCB is asymptotically as efficient as the optimal oracle strategy. We also provide finite-time results when $f$ admits a Taylor expansion of order $2$ in every point. By tuning the number of strata $K$ wisely, it is possible to build an algorithm that is almost as efficient as the optimal oracle strategy.
\end{itemize}

The paper is organized as follows. Section~\ref{s:setting} defines the notations used throughout the paper. Section~\ref{s:asympconv} states the asymptotic lower bound on the mean squared error  of the optimal oracle strategy. In this Section, we also provide an intuition on how the number of samples into each stratum should be linked to the variation of the function in the stratum in order for the mean squared error of the estimate to be small. Section~\ref{s:algo} presents the algorithm LMC-UCB and the first Lemma on how many sub-strata are built in the initial strata. Section~\ref{s:mainresults} finally states that the algorithm LMC-UCB  is almost as efficient as the optimal oracle strategy. We finally conclude the paper. Due to the lack of space, we also provide experiments and proofs in the Supplementary Material.

\vspace{-0.2cm}

\section{Setting}\label{s:setting}

\vspace{-0.1cm}

We consider a function $f: [0,1]^d \rightarrow  \R$. We want to estimate as accurately as possible its integral according to the Lebesgue measure, i.e.~$\int_{[0,1]^d} f(x) dx$. In order to do that, we consider algorithms that stratify the domain in two layers of strata, one more refined than the other. The strata of the refined layer are referred to as sub-strata, and we sample in the sub-strata. We will compare the performances of the algorithms we construct, with the performances of the optimal oracle algorithm that has access to the variations $||\nabla f(x)||_2$ of the function $f$ everywhere in the domain, and is allowed to sample the domain where it wishes.

The first step is to partition the domain $[0,1]^d$ in $K$ measurable \textit{strata}. In this paper, we assume that $K^{1/d}$ is an integer\footnote{This is not restrictive in small dimension, but it may become more constraining for large $d$.}. This enables us to partition, in a natural way, the domain in $K$ \textit{hyper-cubic} strata $(\Omega_k)_{k \leq K}$ of same measure $w_k = \frac{1}{K}$. Each of these strata is a region of the domain $[0,1]^d$, and the $K$ strata form a partition of the domain. 
We write $\mu_{k} = \frac{1}{w_{k}} \int_{\Omega_{k}} f(x) dx$ the mean and $\si_{k}^2 = \frac{1}{w_{k}} \int_{\Omega_{k}} \big(f(x) - \mu_{k}\big)^2 dx$ the variance of a sample of the function $f$ when sampling $f$ at a point chosen at random according to the Lebesgue measure conditioned to stratum $\Omega_k$.

We possess a budget of $n$ samples (which is assumed to be known in advance), which means that we can sample $n$ times the function at any point of $[0,1]^d$. We denote by $\alg$ an algorithm that sequentially allocates the budget by sampling at round $t$ in the stratum indexed by $k_t\in\{1,\dots,K\}$, and returns after all $n$ samples have been used an estimate $\hat \mu_{n}$ of the integral of the function $f$.

We consider strategies that sub-partition each stratum $\Omega_k$ in hyper-cubes of same measure in $\Omega_k$, but of heterogeneous measure among the $\Omega_k$. In this way, the number of sub-strata in each stratum $\Omega_k$ can adapt to the variations $f$ within $\Omega_k$.
The algorithms that we consider return a sub-partition of each stratum $\Omega_k$ in $S_{k}$ sub-strata. We call $\N_k = (\Omega_{k,i})_{i \leq S_{k}}$ the sub-partition of stratum $\Omega_k$. In each of these sub-strata, the algorithm allocates at least one point\footnote{This implies that $\sum_k S_{k} \leq n$.}. We write $X_{k,i}$ the first point sampled uniformly at random in sub-stratum $\Omega_{k,i}$. We write $w_{k,i}$ the measure of the sub-stratum $\Omega_{k,i}$. Let us write $\mu_{k,i} = \frac{1}{w_{k,i}} \int_{\Omega_{k,i}} f(x)dx$ the mean and $\var_{k,i} = \frac{1}{w_{k,i}} \int_{\Omega_{k,i}} \big(f(x) - \mu_{k,i}\big)^2dx$ the variance of a sample of $f$ in sub-stratum $\Omega_{k,i}$ (e.g.~of $X_{k,i} = f(U_{k,i})$ where $U_{k,i} \sim \mathcal U_{\Omega_{k,i}}$).
\\
This class of $2-$layered sampling strategies is rather large. In fact it contains strategies that are similar to low discrepancy strategies, and also to any stratified Monte-Carlo strategy. For example, consider that all $K$ strata are hyper-cubes of same measure $\frac{1}{K}$ and that each stratum $\Omega_k$ is partitioned into $S_{k}$ hyper-rectangles $\Omega_{k,i}$ of minimal diameter and same measure $\frac{1}{KS_{k}}$. If the algorithm allocates one point per sub-stratum, its sampling scheme shares similarities with quasi Monte-Carlo sampling schemes, since the points at which the function is sampled are well spread.

Let us now consider an algorithm that first chooses the sub-partition $(\N_k)_k$ and then allocates deterministically $1$ sample uniformly at random in each sub-stratum $\Omega_{k,i}$. We consider the stratified estimate $\hmu_n = \sum_{k=1}^K \sum_{i=1}^{S_{k}} \frac{w_{k,i}}{S_{k}}  X_{k,i}$ of $\mu$. We have
\begin{equation*}
 \E(\hmu_{n}) = \sum_{k=1}^K \sum_{i=1}^{S_{k}} \frac{w_{k,i}}{S_{k}}   \mu_{k,i} = \sum_{k \leq K}\sum_{i=1}^{S_{k}}  \int_{\Omega_{k,i}} f(x) dx = \int_{[0,1]^d} f(x) dx = \mu,
\end{equation*}
and also
\vspace{-0.3cm}
\begin{equation*}
 \V(\hmu_n) = \sum_{k \leq K} \sum_{i=1}^{S_{k}} (\frac{w_{k,i}}{S_{k}})^2  \E(X_{k,i}-\mu_{k,i})^2 = \sum_{k \leq K} \sum_{i=1}^{S_{k}} \frac{w_{k,i}^2}{S_{k}^2} \var_{k,i}.
\end{equation*}
\vspace{-0.4cm}

For a given algorithm $\alg$ that builds for each stratum $k$ a sub-partition $\N_k = (\Omega_{k,i})_{i \leq S_{k}}$, we call \emph{pseudo-risk} the quantity
\vspace{-0.2cm}
\begin{equation}\label{risk}
 L_n(\alg) = \sum_{k \leq K} \sum_{i=1}^{S_{k}} \frac{w_{k,i}^2}{S_{k}^2} \var_{k,i}.
\end{equation}
Some further insight on this quantity is provided in the paper~\cite{rapp-tech-MC-UCB}.

Consider now the uniform strategy, i.e.~a strategy that divides the domain in $K=n$ hyper-cubic strata. This strategy is a fairly natural, minimax-optimal \textit{static} strategy, on the class of differentiable function defined on $[0,1]^d$, when no information on $f$ is available. We will prove in the next Section that its asymptotic mean squared error is equal to
$$\frac{1}{12} \Big(\int_{[0,1]^d} ||\nabla f(x)||_2^2 dx \Big)  \frac{1}{n^{1 + \frac{2}{d}}}.$$
This quantity is of order $n^{-1-2/d}$, which is smaller, as expected, than $1/n$: this strategy is more efficient than crude Monte-Carlo.

We will also prove in the next Section that the minimum asymptotic mean squared error of an optimal \textit{oracle} strategy (we call it ``oracle'' because it builds the stratification using the information about the variations $||\nabla f(x)||_2$ of $f$ in every point $x$), is larger than
$$\frac{1}{12} \Big(\int_{[0,1]^d}(||\nabla f(x)||_2)^{\frac{d}{d+1}}dx\Big)^{2\frac{(d+1)}{d}} \frac{1}{n^{1 + \frac{2}{d}}}$$
This quantity is always smaller than the asymptotic mean squared error of the Uniform stratified Monte-Carlo strategy, which makes sense since this strategy assumes the knowledge of the variations of $f$ everywhere, and can thus adapt accordingly the number of samples in each region. We define
\begin{equation}\label{eq:sigmabig}
\Sigma = \frac{1}{12} \Big(\int_{[0,1]^d}(||\nabla f(x)||_2)^{\frac{d}{d+1}}dx\Big)^{2\frac{(d+1)}{d}}.
\end{equation}

Given this minimum asymptotic mean squared error of an optimal oracle strategy, we define the pseudo-regret of an algorithm $\alg$ as
\begin{equation}\label{eq:regret}
 R_n(\alg) = L_n(\alg) - \Sigma \frac{1}{n^{1 + \frac{2}{d}}}.
\end{equation}
This pseudo-regret is the difference between the pseudo-risk of the estimate provided by algorithm $\alg$, and the lower-bound on the optimal oracle mean squared error. In other words, this pseudo-regret is the price an adaptive strategy pays for not knowing in advance the function $f$, and thus not having access to its variations. An efficient adaptive strategy should aim at minimizing this gap coming from the lack of informations.

\vspace{-0.2cm}

\section{Discussion on the optimal asymptotic mean squared error}\label{s:asympconv}

\vspace{-0.2cm}

\subsection{Asymptotic lower bound on the mean squared error, and comparison with the Uniform stratified Monte-Carlo}\label{ss:asympconv1}
\vspace{-0.2cm}
A first part of the analysis of the exposed problem consists in finding a good point of comparison for the pseudo-risk. The following Lemma states an asymptotic lower bound on the mean squared error of the optimal oracle sampling strategy.

\begin{lemma}\label{prop:asymbound}
Assume that $f$ is such that $\nabla f$ is continuous and $\int ||\nabla f(x)||_2^2 dx<\infty$. Let $\big((\Omega_{k}^n)_{k \leq n}\big)_n$ be an arbitrary sequence of partitions of $[0,1]^d$ in $n$ strata such that all the strata are hyper-cubes, and such that the maximum diameter of each stratum goes to $0$ as $n \rightarrow +\infty$ (but the strata are allowed to have heterogeneous measures).
 Let $\hat \mu_{n}$ be the stratified estimate of the function for the partition $(\Omega_{k}^n)_{k \leq n}$ when there is one point pulled at random per stratum. Then
 \begin{align*}
 \lim \inf_{n \rightarrow \infty}  n^{1+2/d} \V(\hat \mu_{n})
 &\geq \Sigma.
\end{align*}
\end{lemma}
\vspace{-0.2cm}
The full proof of this Lemma is in the Supplementary Material, Appendix~\ref{proof:asymbound}.



We have also the following equality for the asymptotic mean squared error of the uniform strategy.

\vspace{-0.2cm}

\begin{lemma}\label{prop:unifvar}
Assume that $f$ is such that $\nabla f$ is continuous and $\int ||\nabla f(x)||_2^2 dx<\infty$. For any $n = l^d$ such that $l$ is an integer (and thus such that it is possible to partition the domain in $n$ hyper-cubic strata of same measure), define $\big((\Omega_{k}^n)_{k \leq n}\big)_{n}$ as the sequence of partitions in hyper-cubic strata of same measure $1/n$.
Let $\hat \mu_{n}$ be the stratified estimate of the function for the partition $(\Omega_{k}^n)_{k \leq n}$ when there is one point pulled at random per stratum. Then
 \begin{align*}
 \lim \inf_{n \rightarrow \infty}  n^{1+2/d} \V(\hat \mu_{n})
 &= \frac{1}{12} \Big(\int_{[0,1]^d}||\nabla f(x)||_2^2 dx\Big).
\end{align*}
\end{lemma}
\vspace{-0.2cm}
The proof of this Lemma is substantially similar to the proof of Lemma~\ref{prop:asymbound} in the Supplementary Material, Appendix~\ref{proof:asymbound}. The only difference is that the measure of each stratum $\Omega^n_k$ is $1/n$ and that in Step 2, instead of Fatou's Lemma, the Theorem of dominated convergence is required.

The optimal rate for the mean squared error, which is also the rate of the Uniform stratified Monte-Carlo in Lemma~\ref{prop:unifvar}, is $n^{-1-2/d}$ and is attained with ideas of low discrepancy sampling. The constant can however be improved (with respect to the constant in Lemma~\ref{prop:unifvar}), by adapting to the specific shape of each function. In Lemma~\ref{prop:asymbound}, we exhibit a lower bound for this constant (and without surprises, $\frac{1}{12} \Big(\int_{[0,1]^d}||\nabla f(x)||_2^2 dx\Big) \geq \Sigma$). Our aim is to build an adaptive sampling scheme, also sharing ideas with low discrepancy sampling, that attains this lower-bound.

There is one main restriction in both Lemma: we impose that the sequence of partitions $\big((\Omega_{k}^n)_{k \leq n}\big)_n$ is composed only with strata that have the shape of an hyper-cube. This assumption is in fact reasonable: indeed, if the shape of the strata could be arbitrary, one could take the level sets (or approximate level sets as the number of strata is limited by $n$) as strata, and this would lead to $\lim_{n \rightarrow \infty} \inf_{\Omega}  n^{1+2/d} \V(\hat \mu_{n, \Omega})  =0$. But this is not a fair competition, as the function is unknown, and determining these level sets is actually a much harder problem than integrating the function.
\\
The fact that the strata are hyper-cubes appears, in fact, in the bound. If we had chosen other shapes, e.g.~$l_2$ balls, the constant $\frac{1}{12}$ in front of the bounds in both Lemma would change\footnote{The $\frac{1}{12}$ comes from computing the variance of an uniform random variable on $[0,1]$.}. It is however not possible to make a finite partition in $l_2$ balls of $[0,1]^d$, and we chose hyper-cubes since it is quite easy to stratify $[0,1]^d$ in hyper-cubic strata.

The proof of Lemma~\ref{prop:asymbound} makes the quantity $s^*(x) = \frac{(||\nabla f(x)||_2)^{\frac{d}{d+1}}}{\int_{[0,1]^d} (||\nabla f(u)||_2)^{\frac{d}{d+1}} du}$ appear. This quantity is proposed as ``asymptotic optimal allocation'', i.e.~the asymptotically optimal number of sub-strata one would ideally create in any small sub-stratum centered in $x$. This is however not very useful for building an algorithm. The next Subsection provides an intuition on this matter.
\vspace{-0.2cm}
\subsection{An intuition of a good allocation: Piecewise linear functions}
\vspace{-0.2cm}
In this Subsection, we (i) provide an example where the asymptotic optimal mean squared error is also the optimal mean squared error at finite distance and (ii) provide explicitly what is, in that case, a good allocation. We do that in order to give an intuition for the algorithm that we introduce in the next Section.

We consider a partition in $K$ hyper-cubic strata $\Omega_k$. Let us assume that the function $f$ is affine on all strata $\Omega_k$, i.e.~on stratum $\Omega_k$, we have $f(x) = \Big(\langle \theta_k, x \rangle + \rho_k \Big) \ind{x \in \Omega_k}$. In that case $\mu_{k} = f(a_{k})$ where $a_{k}$ is the center of the stratum $\Omega_k$. We then have:
\begin{align*}
 \si_{k}^2 &= \frac{1}{w_k}\int_{\Omega_{k}} (f(x) - f(a_{k}))^2dx = \frac{1}{w_k}\int_{\Omega_{k}} \Big(\langle \theta_k, (x - a_{k})\rangle \Big)^2dx = \frac{1}{w_k} \Big(\frac{||\theta_k||_2^2}{12} w_k^{1+2/d} \Big) = \frac{||\theta_k||_2^2}{12} w_k^{2/d}.
\end{align*}
We consider also a sub-partition of $\Omega_k$ in $S_{k}$ hyper-cubes of same size (we assume that $S_{k}^{1/d}$ is an integer), and we assume that in each sub-stratum $\Omega_{k,i}$, we sample one point. We also have $\si_{k,i}^2 = \frac{||\theta_k||_2^2}{12} \big(\frac{w_k}{S_{k}}\big)^{2/d}$ for sub-stratum $\Omega_{k,i}$.

For a given $k$ and a given $S_{k}$, all the $\si_{k,i}$ are equals. The pseudo-risk of an algorithm $\alg$ that divides each stratum $\Omega_k$ in $S_{k}$ sub-strata is thus
\begin{equation*}
 L_n(\alg) = \sum_{k \leq K} \sum_{i \leq S_k} \frac{w_k^2 }{S_{k}^2}\frac{||\theta_k||_2^2}{12} \big(\frac{w_k}{S_{k}}\big)^{2/d} = \sum_{k \leq K}  \frac{w_k^{2+2/d} }{S_{k}^{1+2/d}}\frac{||\theta_k||_2^2}{12} = \sum_{k \leq K}  \frac{w_k^{2} }{S_{k}^{1+2/d}} \si_{k}^2.
\end{equation*}
If an unadaptive algorithm $\alg^*$ has access to the variances $\var_{k}$ in the strata, it can choose to allocate the budget in order to minimize the pseudo-risk. After solving the simple optimization problem of minimizing $L_n(\alg)$ with respect to $(S_k)_k$, we deduce that an optimal oracle strategy on this stratification would divide each stratum $k$ in $S_{k}^* = \frac{(w_k \si_{k})^{\frac{d}{d+1}}}{\sum_{i \leq K} (w_i \si_{i})^{\frac{d}{d+1}}}n$ sub-strata\footnote{We deliberately forget about rounding issues in this Subsection. The allocation we provide might not be realizable (e.g.~if $S_k^*$ is not an integer), but plugging it in the bound provides a lower bound on any realizable performance.}. The pseudo-risk for this strategy is then
\begin{equation}\label{BestStrat}
 L_{n,K}(\alg^*) = \frac{\Big(\sum_{k \leq K} (w_k \si_{k})^{\frac{d}{d+1}}\Big)^{2\frac{(d+1)}{d}}}{n^{1+2/d}} = \frac{\Sigma_{K}^{2\frac{(d+1)}{d}}}{n^{1+2/d}},
\end{equation}
where we write $\Sigma_{K} = \sum_{i \leq K} (w_i \si_{i})^{\frac{d}{d+1}}$. We will call in the paper \textit{optimal proportions} the quantities
\begin{equation}\label{eq:lambdak}
\lambda_{K,k} = \frac{(w_k \si_{k})^{\frac{d}{d+1}}}{\sum_{i \leq K} (w_i \si_{i})^{\frac{d}{d+1}}}.
\end{equation}
In the specific case of functions that are piecewise linear, we have $\Sigma_K = \sum_{k \leq K} (w_k \si_{k})^{\frac{d}{d+1}} = \sum_{k \leq K} (w_k \frac{||\theta_k||_2}{2\sqrt{3}} w_k^{1/d})^{\frac{d}{d+1}} = \int_{[0,1]^d} \frac{(||\nabla f(x)||_2)^{\frac{d}{d+1}}}{12^{\frac{d}{2(d+1)}}}dx $. We thus have
\begin{equation}\label{BestStrat2}
 L_{n,K}(\alg^*) =  \Sigma \frac{1}{n^{1 + \frac{2}{d}}}.
\end{equation}

This optimal oracle strategy attains the lower bound in Lemma~\ref{prop:asymbound}. We will thus construct, in the next Section, an algorithm that learns and adapts to the optimal proportions defined in Equation~\ref{eq:lambdak}.

\section{The Algorithm LMC-UCB}\label{s:algo}
\vspace{-0.2cm}

\subsection{Algorithm LMC-UCB}
\vspace{-0.2cm}

We present the algorithm Lipschitz Monte Carlo Upper Confidence Bound ($LMC-UCB$). 
It takes as parameter a partition $(\Omega_k)_{k \leq K}$ in $K \leq n$ hyper-cubic strata of same measure $1/K$ (it is possible since we assume that $\exists l \in \mathbb N / l^d = K$). 
It also takes as parameter an uniform upper bound $L$ on $||\nabla f(x)||_2^2$, and $\delta$, a (small) probability.
The aim of algorithm $LMC-UCB$ is to sub-stratify each stratum $\Omega_k$ in $\lambda_{K,k} = \frac{(w_k\si_k)^{\frac{d}{d+1}}}{\sum_{i=1}^K (w_i\si_i)^{\frac{d}{d+1}}}n$ hyper-cubic sub-strata of same measure and sample one point per sub-stratum. An intuition on why this target is relevant was provided in Section~\ref{s:asympconv}.
\\
Algorithm LMC-UCB starts by sub-stratifying each stratum $\Omega_k$ in $\bar S = \Bigg\lfloor\Big(\big(\frac{n}{K}\big)^{\frac{d}{d+1}}\Big)^{1/d}  \Bigg\rfloor^d$ hyper-cubic strata of same measure. It is possible to do that since by definition, $\bar S^{1/d}$ is an integer. We write this first sub-stratification $\N_k' = (\Omega_{k,i}')_{i\leq \bar S}$. It then pulls one sample per sub-stratum in $\N_k'$ for each $\Omega_k$.
\\
It then sub-stratifies again each stratum $\Omega_k$ using the informations collected. It sub-stratifies each stratum $\Omega_k$ in
\vspace{-0.4cm}
\begin{equation}\label{eq:stra}
S_{k} =\max \Bigg\{ \Bigg\lfloor \Big[ \frac{w_k^{\frac{d}{d+1}}\Big( \hsi_{k,K \bar S} + A(\frac{w_k}{\bar S})^{1/d}\sqrt{\frac{1}{\bar S}} \Big)^{\frac{d}{d+1}}}{ \sum_{i=1}^Kw_i^{\frac{d}{d+1}}\Big( \hsi_{i,K \bar S} + A(\frac{w_i}{\bar S})^{1/d}\sqrt{\frac{1}{\bar S}} \Big)^{\frac{d}{d+1}}} (n - K \bar S) \Big]^{1/d} \Bigg\rfloor^d, \bar S \Bigg\}
\end{equation}
hyper-cubic strata of same measure (see Figure~\ref{f:m-algorithm} for a definition of $A$). It is possible to do that because by definition, $S_{k}^{1/d}$ is an integer. We call this sub-stratification of stratum $\Omega_k$ stratification $\N_k = (\Omega_{k,i})_{i \leq S_k}$.
In the last Equation, we compute the empirical standard deviation in stratum $\Omega_k$ at time $K\bar S$ as
\begin{equation}\label{eq:estim-var2}
 \hsi_{k,K\bar S} = \sqrt{\frac{1}{\bar S-1}\sum_{i=1}^{\bar S}\Big(X_{k,i} - \frac{1}{\bar S}\sum_{j=1}^{\bar S} X_{k,j}\Big)^2}.
\end{equation}

Algorithm LMC-UCB then samples in each sub-stratum $\Omega_{k,i}$ one point. It is possible to do that since, by definition of $S_k$, $\sum_k S_k + K \bar S \leq n$
\\
The algorithm outputs an estimate $\hmu_n$  of the integral of $f$, computed with the first point in each sub-stratum of partition $\N_k$. We present in Figure~\ref{f:m-algorithm} the pseudo-code of algorithm LMC-UCB.
\vspace{-0.4cm}
\begin{figure}[ht]
\bookbox{
\begin{algorithmic}
\STATE \textbf{Input:} Partition $(\Omega_k)_{k \leq K}$, $L$, $\delta$, set $A = 2L\sqrt{d}\sqrt{\log(2K/\delta)}$
\STATE \textbf{Initialize:} $\forall k \leq K$, sample $1$ point in each stratum of partition $\N_k'$
\STATE \textbf{Main algorithm:}
  \STATE Compute $S_k$ for each $k \leq K$
   \STATE Create partition $\N_k$ for each $k \leq K$
  \STATE Sample a point in $\Omega_{k,i} \in \N_k$ for $i \leq S_k$
\STATE \textbf{Output:} Return the estimate $\hmu_n$ computed when taking the first point $X_{k,i}$ in each sub-stratum $\Omega_{k,i}$ of $\N_k$, that is to say $\hmu_n = \sum_{k=1}^K w_k \sum_{i=1}^{S_{k}} \frac{X_{k,i}}{S_{k}}$
\end{algorithmic}}
\vspace{-0.4cm}
\caption{Pseudo-code of LMC-UCB. The definition of  $\N_k'$, $\bar S$, $\N_k$, $\Omega_{k,i}$ and $S_k$ are in the main text.}\label{f:m-algorithm}
\end{figure}
\vspace{-0.4cm}

\vspace{-0.2cm}
\subsection{High probability lower bound on the number of sub-strata of stratum $\Omega_k$}
\vspace{-0.2cm}
We first state an assumption on the function $f$.
\vspace{-0.2cm}
\begin{assumption}\label{ass:bounded}
 The function $f$ is such that $\nabla f$ exists and $\forall x \in [0,1]^d, ||\nabla f(x)||_2^2 \leq L$.
\end{assumption}
\vspace{-0.2cm}
The next Lemma states that with high probability, the number $S_k$ of sub-strata of stratum $\Omega_k$, in which there is at least one point, adjusts ``almost'' to the unknown optimal proportions.
\begin{lemma}\label{lem:nbpulls}
Let Assumption~\ref{ass:bounded} be satisfied and $(\Omega_k)_{k \leq K}$ be a partition in $K$ hyper-cubic strata of same measure. If $n\geq 4K$, then with probability at least $1-\delta$, $\forall k$, the number of sub-strata satisfies
\begin{align*}
S_{k} &\geq \max\Bigg[\lambda_{K,k}\Big[n - 7(L+1)d^{3/2} \sqrt{\log(K/\delta)} (1+\frac{1}{\Sigma_K}) K^{\frac{1}{d+1}} n^{\frac{d}{d+1}}\Big], \bar S\Bigg].
\end{align*}
\end{lemma}
\vspace{-0.4cm}


The proof of this result is in the Supplementary Material (Appendix~\ref{s:m-results}).
\vspace{-0.2cm}
\subsection{Remarks}
\vspace{-0.2cm}

\paragraph{A sampling scheme that shares ideas with quasi Monte-Carlo methods:} Algorithm $LMC-UCB$ \textit{almost} manages to divide each stratum $\Omega_k$ in $\lambda_{K,k}n$ hyper-cubic strata of same measure, each one of them containing at least one sample. It is thus possible to build a learning procedure that, at the same time, estimates the empirical proportions $\lambda_{K,k}$, and allocates the samples proportionally to them.
\vspace{-0.3cm}
\paragraph{The error terms:} There are two reasons why we are not able to divide \textit{exactly} each stratum $\Omega_k$ in $\lambda_{K,k}n$ hyper-cubic strata of same measure. The first reason is that the true proportions $\lambda_{K,k}$ are unknown, and that it is thus necessary to estimate them. The second reason is that we want to build strata that are hyper-cubes of same measure. The number of strata $S_{k}$ needs thus to be such that $S_{k}^{1/d}$ is an integer. We thus also loose efficiency because of rounding issues.

\vspace{-0.3cm}
\section{Main results}\label{s:mainresults}
\vspace{-0.3cm}
\subsection{Asymptotic convergence of algorithm LMC-UCB}
\vspace{-0.2cm}
By just combining the result of Lemma~\ref{prop:asymbound} with the result of Lemma~\ref{lem:nbpulls}, it is possible to show that algorithm LMC-UCB is asymptotically (when $K$ goes to $+\infty$ and $n\geq K$) as efficient as the optimal oracle strategy of Lemma~\ref{prop:asymbound}.
\vspace{-0.2cm}
\begin{theorem}\label{th:asymp.conv}
 Assume that $\nabla f$ is continuous, and that Assumption~\ref{ass:bounded} is satisfied. Let $(\Omega_{k}^n)_{n, k\leq K_n}$ be an arbitrary sequence of partitions such that all the strata are hyper-cubes, such that $4 K_n \leq n$, such that the diameter of each strata goes to $0$, and such that $\lim_{n \rightarrow +\infty}\frac{1}{n}\Big(K_n \big(\log(K_n n^2)\big)^{\frac{d+1}{2}}\Big) = 0$. 
The regret of LMC-UCB with parameter $\delta_n = \frac{1}{n^2}$ on this sequence of partition, where for sequence $(\Omega_{k}^n)_{n, k\leq K_n}$ it disposes of $n$ points, is such that
 \begin{align*}
 \lim_{n \rightarrow \infty} n^{1+2/d}R_n(\alg_{LMC-UCB})
 &= 0.
\end{align*}
\end{theorem}
\vspace{-0.3cm}
The proof of this result is in the Supplementary Material (Appendix~\ref{proof:asymp.conv}).

\vspace{-0.2cm}
\subsection{Under a slightly stronger Assumption}
\vspace{-0.2cm}

We introduce the following Assumption, that is to say that $f$ admits a Taylor expansion of order $2$.
\vspace{-0.2cm}
\begin{assumption}\label{ass:tayexp}
 $f$ admits a Taylor expansion at the second order in any point $a \in [0,1]^d$ and this expansion is such that $\forall x, |f(x) - f(a) - \langle \nabla f, (x-a) \rangle | \leq M ||x-a||_2^2$ where $M$ is a constant.
\end{assumption}
\vspace{-0.2cm}
This is a slightly stronger assumption than Assumption~\ref{ass:bounded}, since it imposes, additional to Assumption~\ref{ass:bounded}, that the variations of $\nabla f(x)$ are uniformly bounded for any $x \in [0,1]^d$.
Assumption~\ref{ass:tayexp} implies Assumption~\ref{ass:bounded} since $\big| ||\nabla f(x)||_2 - ||\nabla f(0)||_2 \big| \leq M ||x-0||_2$, which implies that $ ||\nabla f(x)||_2 \leq ||\nabla f(0)||_2  + M \sqrt{d}$. This implies in particular that we can consider $L = ||\nabla f(0)||_2  + M \sqrt{d}$. We however do not need $M$ to tune the algorithm LMC-UCB, as long as we have access to $L$ (although $M$ appears in the bound of next Theorem).

We can now prove a bound on the pseudo-regret. 
\begin{theorem}\label{th:pi}
 Under Assumptions~\ref{ass:bounded} and~\ref{ass:tayexp}, if $n\geq 4K$, the estimate returned by algorithm $LMC-UCB$ is such that, with probability $1-\delta$, we have
\begin{small}
\begin{align*}
 R_n(\alg_{LMC-UCB})
\leq&  \frac{1}{n^{\frac{d+2}{d}}} \Big[M(L+1)^4  \Big(1 +  \frac{3M d}{\Sigma} \Big)^4  \Big(650d^{3/2} \sqrt{\log(K/\delta)} K^{\frac{1}{d+1}} n^{-\frac{1}{d+1}} + 25d \big(\frac{1}{K}\big)^{\frac{1}{d+1}} \Big)\Big].
\end{align*}
\end{small}
\end{theorem}
%
%
%
%
%
%
%
%
A proof of this result is in the Supplementary Material (Appendix~\ref{proof:thpi})

Now we can choose optimally the number of strata so that we minimize the regret.
\vspace{-0.2cm}
\begin{theorem}\label{th:pi.r}
 Under Assumptions~\ref{ass:bounded} and~\ref{ass:tayexp}, the algorithm $LMC-UCB$ launched on $K_n = \Big\lfloor(\sqrt{n})^{1/d}\Big\rfloor^d$ hyper-cubic strata is such that, with probability $1-\delta$, we have
\begin{align*}
 R_n(\alg_{LMC-UCB})
\leq&   \frac{1}{n^{1+\frac{2}{d} + \frac{1}{2(d+1)}}} \Big[700 M(L+1)^4 d^{3/2} \Big(1 +  \frac{3M d}{\Sigma} \Big)^4  \sqrt{\log(n/\delta)}\Big].
\end{align*}
\end{theorem}
%
%
%
%
%
%
%
%
%
%
%
\vspace{-0.4cm}
\subsection{Discussion}\label{ss:disc}
\vspace{-0.1cm}
\paragraph{Convergence of the algorithm LMC-UCB to the optimal oracle strategy:} When the number of strata $K_n$ grows to infinity, but such that $\lim_{n \rightarrow +\infty}\frac{1}{n}\Big(K_n \big(\log(K_n n^2)\big)^{\frac{d+1}{2}}\Big)= 0$,  the pseudo-regret of algorithm LMC-UCB converges to $0$. It means that this strategy is asymptotically as efficient as (the lower bound on) the optimal oracle strategy. When $f$ admits a Taylor expansion at the first order in every point, it is also possible to obtain a finite-time bound on the pseudo-regret.
\vspace{-0.4cm}
\paragraph{A new sampling scheme:} The algorithm $LMC-UCB$ samples the points in a way that takes advantage of both stratified sampling and quasi Monte-Carlo. Indeed, LMC-UCB is designed to cumulate (i) the advantages of quasi Monte-Carlo by spreading the samples in the domain and (ii) the advantages of stratified, adaptive sampling by allocating more samples where the function has larger variations. 
 For these reasons, this technique is very efficient on differentiable functions. We illustrate this assertion by numerical experiments in the Supplementary Material (Appendix~\ref{app:exp}).
\vspace{-0.4cm}
\paragraph{In high dimension:} The bound on the pseudo-regret in Theorem~\ref{th:pi.r} is of order $n^{-1 - \frac{2}{d}} \times poly(d) n^{-\frac{1}{2(d+1)}}$. In order for the pseudo-regret to be negligible when compared to the optimal oracle mean squared error of the estimate (which is of order $n^{-1 - \frac{2}{d}}$) it is necessary that $poly(d)n^{-\frac{1}{2(d+1)}}$ is negligible compared to $1$. In particular, this says that $n$ should scale exponentially with the dimension $d$. This is unavoidable, since stratified sampling shrinks the approximation error to the asymptotic oracle only if the diameter of each stratum is small, i.e.~if the space is stratified in every direction (and thus if $n$ is exponential with $d$). However Uniform stratified Monte-Carlo, also for the same reasons, shares this problem\footnote{When $d$ is very large and $n$ is not exponential in $d$, then second order terms, depending on the dimension, take over the bound in Lemma~\ref{prop:unifvar} (which is an asymptotic bound) and $poly(d)$ appears in these negligible terms.}.
\\
We emphasize however the fact that a (slightly modified) version of our algorithm is more efficient than crude Monte-Carlo, up to a negligible term \textit{that depends only of $poly(\log(d))$}. The bound in Lemma~\ref{lem:nbpulls} depends of $poly(d)$ only because of rounding issues, coming from the fact that we aim at dividing each stratum $\Omega_k$ in hyper-cubic sub-strata. The whole budget is thus not completely used, and only $\sum_k S_k + K \bar S$ samples are collected. 
 By modifying LMC-UCB so that it allocates the remaining budget uniformly at random on the domain, it is possible to prove that the (modified) algorithm is always at least as efficient as crude Monte-Carlo.
\vspace{-0.5cm}
\section*{Conclusion}
\vspace{-0.4cm}
This work provides an adaptive method for estimating the integral of a differentiable function $f$.
\\
We first proposed a benchmark for measuring efficiency: we proved that the asymptotic mean squared error of the estimate outputted by the optimal oracle strategy is lower bounded by $\Sigma\frac{1}{n^{1+2/d}}$.
\\
We then proposed an algorithm called LMC-UCB, which manages to learn the amplitude of the variations of $f$, to sample more points where theses variations are larger, and to spread these points in a way that is related to quasi Monte-Carlo sampling schemes. 
We proved that algorithm LMC-UCB is asymptotically as efficient as the optimal, oracle strategy. Under the assumption that $f$ admits a Taylor expansion in each point, we provide also a finite time bound for the pseudo-regret of algorithm LMC-UCB. 
We summarize in Table~\ref{f:rates} the rates and finite-time bounds for crude Monte-Carlo, Uniform stratified Monte-Carlo and LMC-UCB.
\vspace{-0.3cm}
\begin{table}[ht]
\begin{center}
\begin{tabular}{|c||ccc|}
  \hline
&&\textbf{Pseudo-Risk:}&\\
\textbf{Sampling schemes} & Rate & Asymptotic constant & + Finite-time bound\\
\hline
\hline
  Crude MC & $\frac{1}{n}$&  $\int_{[0,1]^d} \big(f(x) - \int_{[0,1]^d}f(u) du\big)^2 dx$ & $+0$\\
  \hline
  Uniform stratified MC & $\frac{1}{n^{1 + \frac{2}{d}}}$& $\frac{1}{12} \Big(\int_{[0,1]^d} ||\nabla f(x)||_2^2 dx \Big)$ & $+ O(\frac{d}{n^{1 + \frac{2}{d} + \frac{1}{2d}}})$ \\
  \hline
  LMC-UCB & $\frac{1}{n^{1 + \frac{2}{d}}} $& $\frac{1}{12} \Big(\int_{[0,1]^d}(||\nabla f(x)||_2)^{\frac{d}{d+1}}dx\Big)^{2\frac{(d+1)}{d}}$ & $+ O(\frac{d^{\frac{11}{2}}}{n^{1 + \frac{2}{d} +\frac{1}{2(d+1)}}})$ \\
  \hline
\end{tabular}
\end{center}
\vspace{-0.4cm}
\caption{Rate of convergence plus finite time bounds for Crude Monte-Carlo, Uniform stratified Monte Carlo (see Lemma~\ref{prop:unifvar}) and LMC-UCB (see Theorems~\ref{th:asymp.conv} and~\ref{th:pi.r}).}\label{f:rates}
\end{table}
\vspace{-0.2cm}
An interesting extension of this work would be to adapt it to $\alpha-$H\"older functions that admit a Riemann-Liouville derivative of order $\alpha$. We believe that similar results could be obtained, with an optimal constant and a rate of order $n^{1+2\alpha/d}$.

\vspace{-0.3cm}

\paragraph{Acknowledgements} This research was partially supported by Nord-Pas-de-Calais Regional Council, French ANR EXPLO-RA (ANR-08-COSI-004), the European Community’s Seventh Framework Programme (FP7/2007-2013) under grant agreement 270327 (project CompLACS), and by Pascal-2.

\newpage
\appendix

\begin{huge}
 Supplementary Material for paper
\\
Adaptive Stratified Sampling for Monte-Carlo integration of Differentiable functions
\end{huge}

\section{Numerical Experiments}\label{app:exp}

We provide some experiments illustrating how LMC-UCB works, and compare its efficiency to that of crude Monte-Carlo and Uniform stratified Monte-Carlo.

\begin{center}
\begin{figure*}[h]
\begin{center}
\includegraphics[width=14cm]{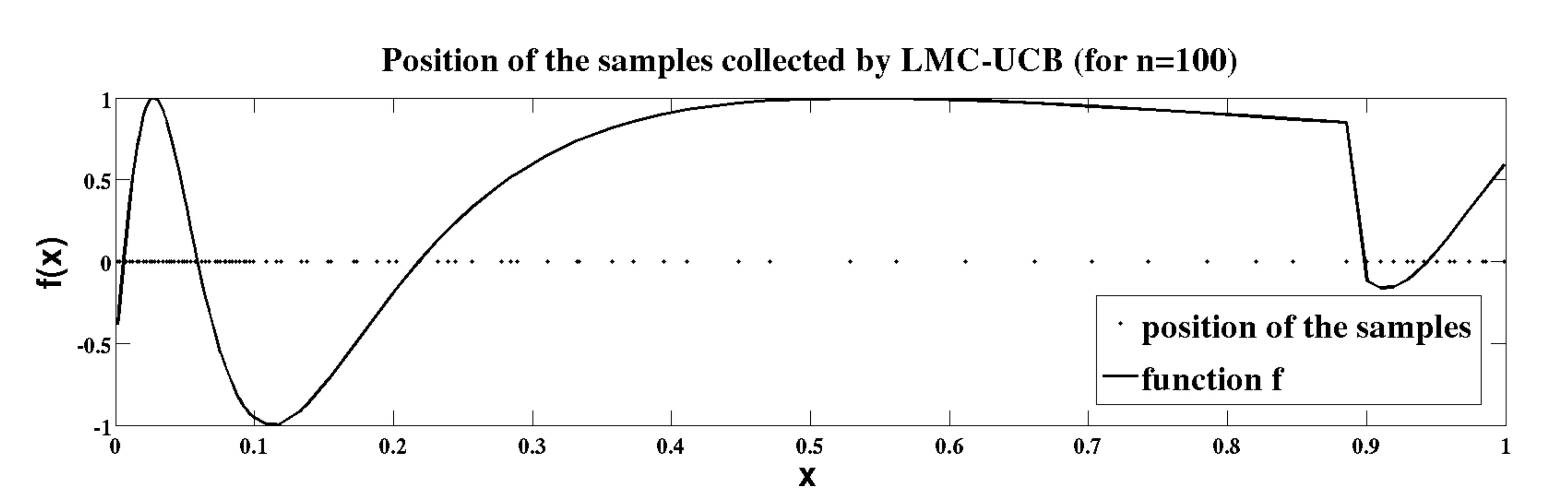}
\caption{Position of the samples collected by LMC-UCB.} \label{fig:exp}
\end{center}
\end{figure*}
\end{center}

We first illustrate on an example, in Figure~\ref{fig:exp}, the sampling scheme. We have launched LMC-UCB on the function displayed in Figure~\ref{fig:exp} (i.e.~$f(x) = \sin(1/(x+0.1)) + \ind{x >0.9} \sin(1/(x-0.7))$). We chose this function since its variations are quite heterogeneous in the domain $[0,1]$.  We considered a budget of $n=100$, and took as parameter $A = 10$. $K_n$ and $\bar S$ are defined as in Figure~\ref{f:m-algorithm}.
We observe that, as expected, the algorithm allocates more points in parts of the domain where the function has larger variations and, additional to that, it spreads the points on the domain so that every region is covered (in a similar spirit to what low-discrepancy schemes would do).

\begin{center}
\begin{figure*}[h]
\begin{minipage}{7cm}
\hspace{-0.83cm}
\includegraphics[width=14cm]{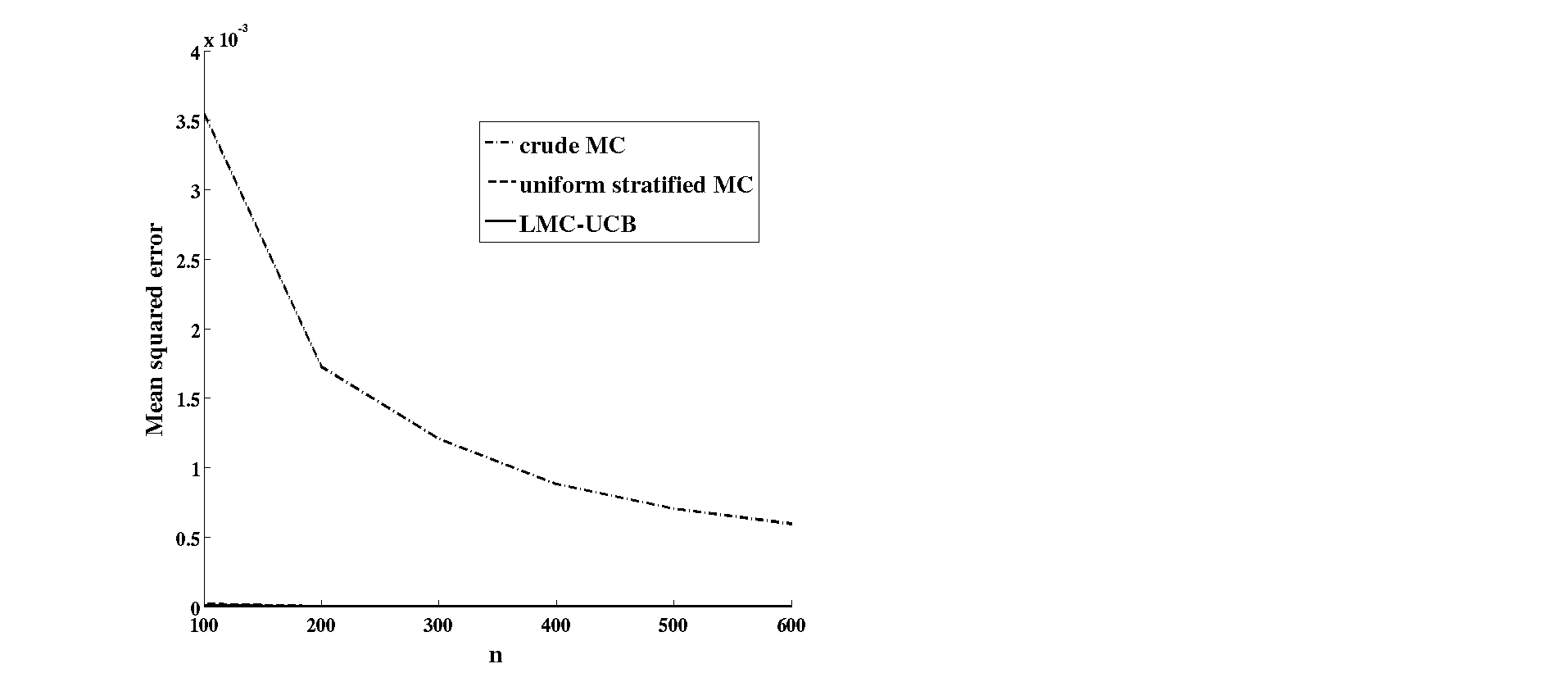}
\caption{Mean squared error w.r.t.~the integral of $f$ of crude Monte-Carlo, uniform stratified Monte-Carlo and LMC-UCB, in function of the budget $n$. Since crude Monte-Carlo is approximately $100$ times less efficient than the two other strategies, their curves are shrinked and not very visible.} \label{fig:exp2}
\end{minipage}
\hspace{0.2cm}
\begin{minipage}{7cm}
\hspace{-1cm}
\includegraphics[width=14cm]{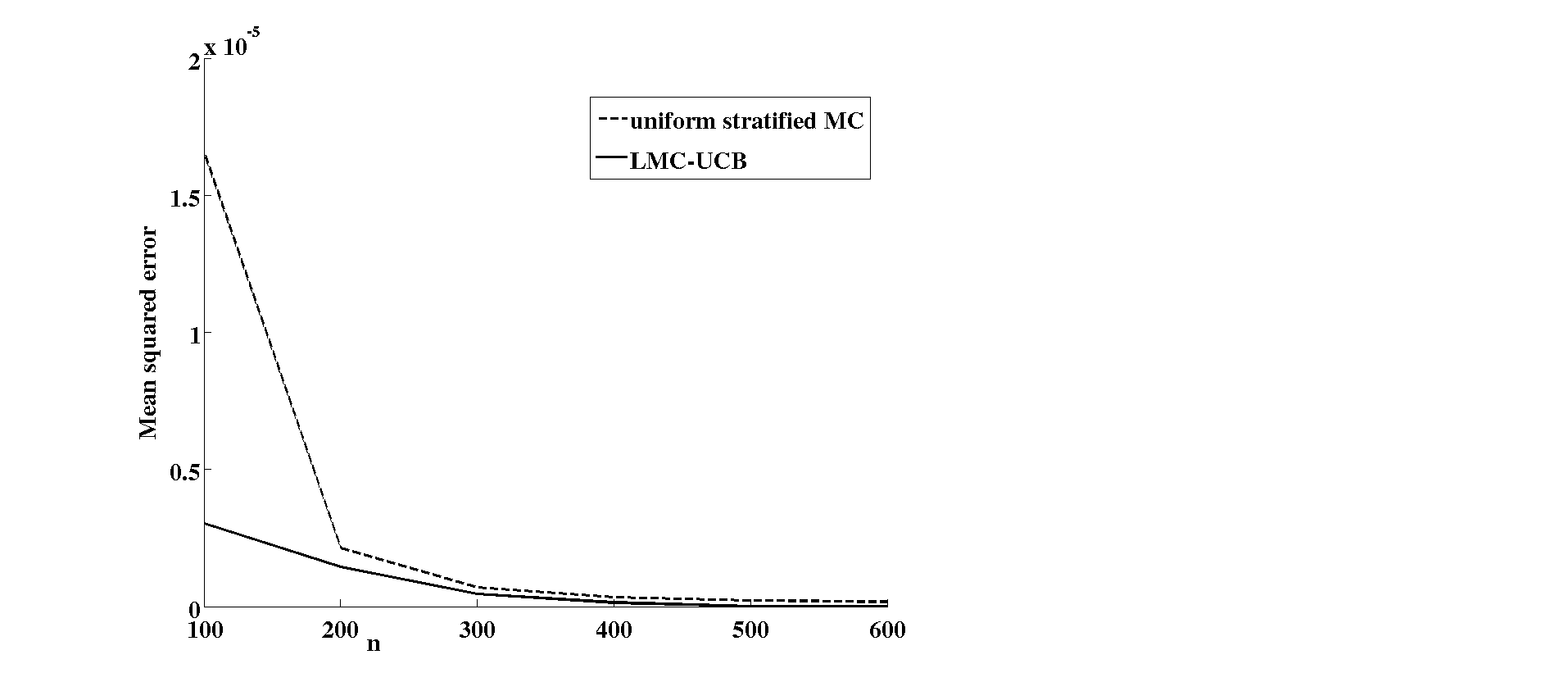}
\caption{Zoom on the mean squared error w.r.t.~the integral of $f$ of uniform stratified Monte-Carlo and LMC-UCB, in function of the budget $n$.} \label{fig:exp3}
\end{minipage}
\end{figure*}
\end{center}

We also compare, for this function, the mean squared error of crude Monte-Carlo, uniform stratified Monte-Carlo and LMC-UCB, for different values of $n$. We average the mean squared error of the estimate returned by each method on $10000$ runs. We have the following performances for each method (displayed in Figures~\ref{fig:exp2} and~\ref{fig:exp3}).

As expected, the mean square error decreases faster than $1/n$ for uniform stratified Monte-Carlo and LMC-UCB. These methods are also more efficient than crude Monte-Carlo (up to $100$ times more efficient on this function), which makes sense since the function that we integrate is differentiable (and then the rate for LMC-UCB and Uniform stratified Monte-Carlo is of order $O(n^{-1-2/d})$). The gain in efficiency when compared to crude Monte-Carlo however decreases with the dimension, as explained in Subsection~\ref{ss:disc}. We observe that LMC-UCB is more efficient than uniform stratified Monte-Carlo, which is a minimax-optimal strategy in the class of non-adaptive strategies.

\section{Poof of Lemma~\ref{prop:asymbound}}\label{proof:asymbound}

\paragraph{Step 0: Decomposition of the variance}
Let $\Omega = (\Omega_{k}^n)_{0<n<+\infty,k\leq n}$ be a sequence of partitions of $[0,1]^d$ in $n$ hyper-cubic strata such that the maximum diameter of the strata in the partitions converges to $0$ when $n$ goes to infinity. In each of those strata, there is a point.

Let $n$ be the number of points, and $k\leq n$ be an index. Let $a_{n,k}$ be a point of the stratum $\Omega_{k}^n$. Let us assume that $f$ is differentiable, that it's derivative $\nabla f$ is continuous, and let us also assume that $||\nabla f (u)||_2^2 = \sum_{i=1}^d \big(\frac{\partial f(u)}{\partial x_i}\big)^2$ is such that $\int ||\nabla f(x)||_2^2 dx$ is bounded. In that case, $\forall x \in \Omega_{k}^n$, there exists $u_{n,k,x} \in \Omega_{k}^n$ such that we have $f(x) - f(a_k) = \langle \nabla f(u_{n,k,x}), x-a_{n,k} \rangle$ (intermediate values theorem). Note also that we have in that case $\mu_{n,k} = f(a_{n,k}) + \frac{1}{w_{n,k}}\int_{\Omega_{k}^n} \langle \nabla f(u_{n,k,x}), x-a_{n,k} \rangle dx$ where $a_{n,k}$ is the center of the stratum $\Omega_{k}^n$. We thus have:
\begin{align*}
 \si_{n,k}^2 =& \frac{1}{w_{n,k}}\int_{\Omega_{k}^n} (f(x) - f(a_{n,k}))^2dx\\
=& \frac{1}{w_{n,k}}\int_{\Omega_{k}^n} \Big(\langle \nabla f(u_{n,k,x}), x-a_{n,k} \rangle -   \frac{1}{w_{n,k}}\int_{\Omega_{k}^n} \langle \nabla f(u_{n,k,y}), y-a_{n,k} \rangle dy\Big)^2dx\\ 
=& \frac{1}{w_{n,k}}\int_{\Omega_{k}^n} \Big(\langle \nabla f(u_{n,k,x}), x-a_{n,k} \rangle\Big)^2dx -   \Big(\frac{1}{w_{n,k}}\int_{\Omega_{k}^n} \langle \nabla f(u_{n,k,y}), y-a_{n,k} \rangle dy\Big)^2\\
=& \frac{1}{w_{n,k}}\int_{[0,1]^d} \Big(\langle \nabla f(u_{n,k,x})\ind{\Omega_{k}}, (x-a_{n,k})\ind{\Omega_{k}^n} \rangle\Big)^2dx\\
&-   \Big(\frac{1}{w_{n,k}}\int_{[0,1]^d} \langle \nabla f(u_{n,k,y})\ind{\Omega_{k}^n}, (y-a_{n,k})\ind{\Omega_{k}^n} \rangle dy\Big)^2.
\end{align*}

\paragraph{Step 1: Convergence of $\si_k$ when the size of the strata goes to $0$}
Let $x\in [0,1]^d$. Note that as as $(\Omega_{k}^n)_{k\leq n}$ is a partition, there is a $k_{n,x}$ such that $x \in \Omega^n_{ k_{n,x}}$.

 Note first that $\nabla f$ is continuous. This means that $\forall \epsilon, \exists \eta / \forall y \in \B_2(x,\eta), ||\nabla f(y) - \nabla f(x)||_2 \leq \epsilon$. Let $\epsilon>0$ and $n$ sufficiently large (any $n$ larger than some given horizon $n'$), the maximum diameter of $\Omega^n_{ k_{n,x}}$ is smaller than $\eta$.
Let $y\in \Omega^n_{k_{n,x}} $. As $u_{n,k_{n,x},y} \in \Omega^n_{ k_{n,x}}$, we know that $||u_{n,k_{n,x},y}-x||\leq \eta$ and that we thus have $||\nabla f(u_{n,k_{n,x},y}) - \nabla f(x)||_2 \leq \epsilon$. This means that $\nabla f(u_{n,k_{n,x},y} )$ converges point-wise to $\nabla f(x)$.

Note also that we have by Cauchy-Schwartz that
 \begin{align*}
\frac{1}{w_{n,k_{n,x}}^{2/d}} \Big(\langle \nabla f(u_{n,k_{n,x},y} ), (y-a_{n,k_{n,x}}) \rangle\Big)^2 \ind{\Omega^n_{k_{n,x}}} &\leq \frac{1}{w_{n,k_{n,x}}^{2/d}} ||\nabla f(u_{n',k_{n',x},y})||_2^2  ||y-a_{n,k_{n,x}}||_2^2 \ind{\Omega^n_{k_{n,x}}}\\
&\leq d||\nabla f(u_{n,k_{n,x},y})||_2^2 \leq dL^2.
\end{align*}

As $\nabla f(u_{n,k_{n,x},y} )$ converges point-wise with $n$ to $\nabla f(x)$, and as $\frac{1}{w_{n,k_{n,x}}^{2/d}} \Big(\langle \nabla f(u_{n,k_{n,x},y} ), (y-a_{n,k_{n,x}}) \rangle\Big)^2
\leq dL^2$, we have by the Theorem of Dominated convergence, that
\begin{align*}
 &\lim_{n \rightarrow +\infty}\frac{1}{w_{n,k_{n,x}}^{1+2/d}}\int_{[0,1]^d} \Big(\langle \nabla f(u_{n,k_{n,x},y} ), (y-a_{n,k_{n,x}}) \rangle\Big)^2 \ind{\Omega^n_{k_{n,x}}}dy\\
&\lim_{n \rightarrow +\infty}\frac{1}{w_{n,k_{n,x}}^{1+2/d}}\int_{[0,1]^d} \Big(\langle \lim_{n \rightarrow +\infty} \nabla f(u_{n,k_{n,x},y} ), (y-a_{n,k_{n,x}}) \rangle\Big)^2 \ind{\Omega^n_{k_{n,x}}}dy\\ 
 &\lim_{n \rightarrow +\infty}\frac{1}{w_{n,k_{n,x}}^{1+2/d}}\int_{[0,1]^d} \Big(\langle \nabla f(x), (y-a_{n,k_{n,x}}) \rangle\Big)^2 \ind{\Omega^n_{k_{n,x}}}dy\\ 
&= \lim_{n \rightarrow +\infty}\frac{1}{w_{n,k_{n,x}}^{1+2/d}} \frac{||\nabla  f(x)||_2^2 w_{n,k_{n,x}}^{1 + 2/d}}{12}\\
&= \frac{||\nabla f(x)||_2^2}{12}.
\end{align*}
In the same way, we have that
\begin{align*}
 &\lim_{n \rightarrow +\infty}\frac{1}{w_{n,k_{n,x}}^{1+2/d}}\Big(\int_{[0,1]^d} \Big(\langle \nabla f(u_{n,k_{n,x},y}), (y-a_{n,k_{n,x}}) \rangle \ind{\Omega^n_{k_{n,x}}} dy\Big)^2\\
&\lim_{n \rightarrow +\infty}\frac{1}{w_{n,k_{n,x}}^{1+2/d}}\Big( \int_{[0,1]^d} \langle \lim_{n \rightarrow +\infty} \nabla f(u_{n,k_{n,x},y}), (y-a_{n,k_{n,x}}) \rangle \ind{\Omega^n_{k_{n,x}}} dy\Big)^2\\ 
 &\lim_{n \rightarrow +\infty}\frac{1}{w_{n,k_{n,x}}^{1+2/d}}\Big( \int_{[0,1]^d} \langle \nabla f(x), (y-a_{n,k_{n,x}}) \rangle \ind{\Omega^n_{k_{n,x}}} dy \Big)^2\\ 
&= \lim_{n \rightarrow +\infty}\frac{1}{w_{n,k_{n,x}}^{1+2/d}} w_{n,k_{n,x}}^{1 + 2/d} \Big(a_{n,k_{n,x}} -a_{n,k_{n,x}} \Big)\\
&= 0.
\end{align*}

Let us call $g_{n,\Omega}(x) =  \sum_{k=1}^n \frac{\si_{n,k}^2}{w_{n,k}^{1/2d}} \ind{\Omega^n_{k}}(x) = \frac{\si_{n,k_{n,x}}^2}{w_{n,k_{n,x}}^{1/2d}}$. The last two inequalities prove, $\forall x$, point-wise convergence of $g_{n,\Omega}(x)$ to $\frac{||\nabla f(x)||_2^2}{12}$: 


\paragraph{Step 2: Optimal allocation and minimum for the asymptotic variance}
There is one point pulled at random per stratum. The variance of the estimate given by such an allocation is
\begin{align*}
 \sum_{k=1}^n w_{n,k}^2 \si_{n,k}^2 = \sum_{k=1}^n w_{n,k}\times w_{n,k}^{1+2/d}\times \frac{\si_{n,k}^2}{w_{n,k}^{2/d}}.
\end{align*}

Define $s_{n,\Omega}(x) = \sum_{k=1}^n \frac{1}{nw_{n,k}} \ind{\Omega^n_{k}}(x)$. Note first that
\begin{align*}
 1 = \frac{1}{n} \sum_{k=1}^n 1 = \int_{[0,1]^d} s_{n,\Omega}(x) dx,
\end{align*}
and that
\begin{align*}
 s_{n,\Omega}(x) >0.
\end{align*}

One has also for the variance of the estimate that
\begin{align*}
 \sum_{k=1}^n w_{n,k}^2 \si_{n,k}^2 = \frac{1}{n^{1+2/d}} \int_{[0,1]^d} g_{n,\Omega}(x) \frac{1}{s_{n,\Omega}(x)^{1+2/d}} dx.
\end{align*}

By using the result of the previous step, one has (for every sequence $\Omega$ where the diameter of the strata converge uniformly to $0$), point-wise convergence of $g_{n,\Omega}(x)$ to $\frac{||\nabla f(x)||_2^2}{12}$ when $n$ goes to infinity.
\\
This leads to, by using Fatou's Lemma
\begin{align*}
&\lim \inf_{n \rightarrow +\infty} \int_{[0,1]^d} g_{n,\Omega}(x) \frac{1}{s_{n,\Omega}(x)^{1+2/d}} dx\\ 
&\geq \int_{[0,1]^d} \lim \inf_{n \rightarrow +\infty} \Big( g_{n,\Omega}(x)  \frac{1}{s_{n,\Omega}(x)^{1+2/d}} \Big) dx\\
&\geq    \int_{[0,1]^d}  \inf_{s: s\geq 0, \int s =1} \frac{||\nabla f(x)||_2^2}{12}  \frac{1}{s(x)^{1+2/d}} dx.
\end{align*}

One thus wants then to find the function $s(x)$ that minimizes this limit. One thus wants to solve in each point $x$ the program $\inf_s \frac{||\nabla f(x)||_2^2}{12} \frac{1}{s(x)^{1+2/d}}$ such that $s\geq 0$ and $\int_{[0,1]^d} s(x) dx=1$.
\\
The solution (by just writing Lagragian) is
\begin{align*}
s^*(x) = \frac{(||\nabla f(x)||_2)^{\frac{d}{d+1}}}{\int_{[0,1]^d} (||\nabla f(u)||_2)^{\frac{d}{d+1}} du}.
\end{align*}

By plugging it in the bound, one obtains
\begin{align*}
&\lim\inf_{n \rightarrow +\infty}  \int_{[0,1]^d} g_{n,\Omega}(x) \frac{1}{s_{n,\Omega}(x)^{1+2/d}} dx\\ 
&\geq   \frac{\Big(\int_{[0,1]^d}(||\nabla f(x)||_2)^{\frac{d}{d+1}}dx\Big)^{2\frac{(d+1)}{d}}}{12}.
\end{align*}

Note that the previous result holds for any sequence of partitions $(\Omega_n)_n$ where the diameter of each stratum converges uniformly to $0$.
One finally has, using that, that the minimum possible asymptotic variance is bounded by
\begin{align*}
 \lim_{n \rightarrow +\infty} \inf_{\Omega} n^{1+2/d} \sum_{k=1}^n w_{n,k}^2 \si_{n,k}^2 \geq \frac{\Big(\int_{[0,1]^d}(||\nabla f(x)||_2)^{\frac{d}{d+1}}dx\Big)^{2\frac{(d+1)}{d}}}{12},
\end{align*}
and we thus obtain the desired result.

\section{Proof of Lemmas~\ref{lem:nbpulls}}\label{s:m-results}

\paragraph{Upper bound on the standard deviation:} The upper confidence bounds $B_{k,t}$ used in the MC-UCB algorithm is an elaboration in the specific case of Lipschitz function on Theorem~10~in~\cite{maurer2009empirical} (a variant of this result is also reported in \cite{audibert2009exploration}). We state here  a main Lemma.

\begin{lemma}\label{l:event-B-AS}
Assume that the function $f$ from which the data is collected is differentiable, and that $||\nabla f(x)||_2$ is bounded by $L$, and $n\geq 2$. Define the following event
%
\begin{equation}\label{eq:ucb-maurer-event-sub1}
\xi = \xi_{K,n}(\delta) = \bigcap_{1\leq k\leq K,}\left\lbrace \Bigg|\sqrt{\frac{1}{\bar S-1}\sum_{i=1}^{\bar S}\Big(X_{k,i} - \frac{1}{\bar S}\sum_{j=1}^{\bar S} X_{k,j}\Big)^2} - \si_k\Bigg| \leq 2L \sqrt{d}(\frac{w_k}{\bar S})^{1/d}\sqrt{\frac{\log(2K/\delta)}{\bar S}} \right\rbrace.
\end{equation}

The probability of $\xi$ is bounded by $1-\delta$.
\end{lemma}
%

Note that the first term in the absolute value in Equation~\ref{eq:ucb-maurer-event-sub1} is the empirical standard deviation of arm $k$ computed as in Equation~\ref{eq:estim-var2} for $t$ samples. The event $\xi$ plays an important role in the proofs of this section and a number of statements will be proved on this event.

We now provide the proof of Lemma~\ref{l:event-B-AS}.

Let us assume that $f$ is such that $||\nabla f||_2 \leq L$. Let us consider a small box $\Omega_w$ of size $w$ and such that $\Omega_w = \prod_{i=1}^d [a_i - \frac{w^{1/d}}{2}, a_i + \frac{w^{1/d}}{2}]$.
As $||\nabla f||_2 \leq L$, we know that $|f(x) - \frac{1}{w} \int_{\Omega_w} f(u)du| \leq L \sqrt{d}w^{1/d}$.
\\
If $U$ is a random variable on $\Omega_w$ and $X = f(U)$, then
\begin{equation*}
|X - \mu| \leq  L\sqrt{d}w^{1/d},
\end{equation*}
where $\mu = \frac{1}{w} \int_{\Omega_w} f(u)du$.

Note first that for algorithm LMC-UCB, the $\bar S$ first samples are each sampled in an hypercube of measure $\frac{w_k}{\bar S}$, and all of those hypercubes form a partition of the domain.
\\
Using a large deviation bound on the variance, e.g.~the one in \cite{maurer2009empirical}, we can deduce that with probability $1-2\delta$
\begin{align*}
 |\sqrt{\frac{1}{\bar S-1}\sum_{i=1}^{\bar S}\Big(X_{k,i} - \frac{1}{\bar S}\sum_{j=1}^{\bar S} X_{k,j}\Big)^2} - \si_k| \leq b\sqrt{\frac{2\log(1/\delta)}{\bar S-1}},
\end{align*}
where $b$ is a bound on the random variables $X_i - \mu_i$. One gets because $|X_{k,i} - \mu_{k,i}| \leq  \sqrt{d}L(\frac{w_k}{t})^{1/d}$ (where $\mu_{k,i}$ is the mean of the function on the hypercube where point $X_{k,i}$ is sampled and because $t \geq 2$
\begin{align*}
 |\sqrt{\frac{1}{\bar S-1}\sum_{i=1}^{\bar S}\Big(X_{k,i} - \frac{1}{\bar S}\sum_{j=1}^{\bar S} X_{k,j}\Big)^2} - \si_k| \leq  2L\sqrt{d}(\frac{w_k}{\bar S})^{1/d}\sqrt{\frac{\log (1/\delta)}{\bar S}}.
\end{align*}
Then by doing a simple union bound on $(k,t)$, we obtain the result.

The following Corollary holds.
\begin{corollary}
 On the event $\xi$, $\forall k \leq K$,
$$|\hsi_{k,K \bar S} - \si_k| \leq 2L\sqrt{d}\sqrt{\log(2K/\delta)}\frac{w_k^{1/d}}{\bar S^{\frac{d+2}{2d}}}$$
\end{corollary}
By concavity, we also have the following Corollary.
\begin{corollary}\label{l:bernstein-var-sub}
 On the event $\xi$, there is $\forall k \leq K$ that
$$|\hsi_{k,K \bar S}^{\frac{d}{d+1}} - \si_k^{\frac{d}{d+1}}| \leq A\frac{w_k^{\frac{1}{d+1}}}{\bar S^{\frac{d+2}{2(d+1)}}},$$
where $A = (2L\sqrt{d}\sqrt{\log(2K/\delta)})^{\frac{d}{d+1}}$.

\end{corollary}

\paragraph{The number of sub-strata}

Let $k$ be an index. Let us call $C_k = \frac{w_k^{\frac{d}{d+1}}\Big( \hsi_{k,K \bar S} + A(\frac{w_k}{\bar S})^{1/d}\sqrt{\frac{1}{\bar S}} \Big)^{\frac{d}{d+1}}}{ \sum_{i=1}^Kw_i^{\frac{d}{d+1}}\Big( \hsi_{i,K\bar S} + A(\frac{w_i}{\bar S})^{1/d}\sqrt{\frac{1}{\bar S}} \Big)^{\frac{d}{d+1}}} (n - K\bar S)$.

Stratum $\Omega_k$ is subdivided in $S_{k} = \max\Big[\bar S , \lfloor C_k^{1/d} \rfloor^d\Big]$ substrata, composing the sub-partition $\N_k$.

Note first that $\sum_{k=1}^K S_{k} \leq n$ as $\sum_{k=1}^K C_k = n-K \bar S$. As the samples are always picked in sub-strata that have the less points, it ensures that there is at least one point per sub-stratum.

On $\xi$, we have because of Corollary~\ref{l:bernstein-var-sub} that
\begin{align*}
 C_k &\geq \frac{w_k^{\frac{d}{d+1}} \si_{k}^{\frac{d}{d+1}}}{ \sum_{i=1}^K w_i^{\frac{d}{d+1}} \Big( \si_{i}^{\frac{d}{d+1}} + 2A \frac{w_i^{\frac{1}{d+1}}}{\bar S^{\frac{d+2}{2(d+1)}}}  \Big)} (n - K\bar S)\\
&\geq \frac{w_k^{\frac{d}{d+1}} \si_{k}^{\frac{d}{d+1}}}{ \Sigma_K + 2A \frac{1}{\bar S^{\frac{d+2}{2(d+1)}}} } (n - K\bar S)\\
&\geq \lambda_{K,k} (n - K\bar S) \Big(1 - \frac{2A}{\Sigma_K \bar S^{\frac{d+2}{2(d+1)}}}\Big)\\
&\geq \lambda_{K,k} \Big(n - K\bar S - \frac{2A n}{\Sigma_K \bar S^{\frac{d+2}{2(d+1)}}}\Big).
\end{align*}

Using the fact that $\big(\frac{n}{K}\big)^{\frac{d}{d+1}} \geq \bar S \geq \Big(\big(\frac{n}{K}\big)^{\frac{1}{d+1}} - 1\Big)^d \geq \big(\frac{n}{K}\big)^{\frac{d}{d+1}} - d \big(\frac{n}{K}\big)^{\frac{d-1}{d+1}}$ in the last Equation,
\begin{align}
 C_k
&\geq \lambda_{K,k} \Big(n -  K \big(\frac{n}{K}\big)^{\frac{d}{d+1}} - \frac{2An}{\Sigma_K} \big(\frac{K}{n}\big)^{\frac{d}{d+1} \times \frac{d+2}{2(d+1)}} \big(1 + d(\frac{K}{n})^{\frac{1}{d+1}} \big)^{\frac{d+2}{2(d+1)}}\Big) \nonumber\\
&\geq \lambda_{K,k} \Big(n -  K^{\frac{1}{d+1}} n^{\frac{d}{d+1}} - \frac{2An^{\frac{1}{2} + \frac{1}{(d+1)^2}}}{\Sigma_K} K^{\frac{d(d+2)}{2(d+1)^2} }  \big(1 + [d(\frac{K}{n})^{\frac{1}{d+1}}]^{\frac{d+2}{2(d+1)}} \big)\Big) \nonumber\\
&\geq \lambda_{K,k} \Big(n - (1 + 2 \frac{A}{\Sigma_K} + d(\frac{K}{n})^{\frac{d+2}{2(d+1)^2}}) K^{\frac{1}{d+1}} n^{\frac{d}{d+1}} \Big), \label{eq:ck}
\end{align}
where the last line comes from the fact that $n \geq K$.

We also have
\begin{align*}
C_k - \lfloor C_k^{1/d} \rfloor^{d} \leq C_k - (C_k^{1/d}-1)^{d} = C_k \big( 1 - (1- \frac{1}{C_k^{1/d}})^d\big) \leq d C_k^{\frac{d-1}{d}}.
\end{align*}

From the last Equation, the definition of $S_{k}$ and Equation~\ref{eq:ck} we deduce that (rounding issues)
\begin{align*}
 S_{k} &\geq \max\Big[\bar S , C_k\big(1 - \frac{d}{C_k^{1/d}}\big) \Big]\\
&\geq \max\Big[\bar S , C_k\big(1 - \frac{d}{(\bar S)^{1/d}}\big) \Big]\\
&\geq \max\Big[\bar S , \lambda_{K,k} \Big(n - (1 + 2 \frac{A}{\Sigma_K} + d(\frac{K}{n})^{\frac{d+2}{2(d+1)^2}})  K^{\frac{1}{d+1}} n^{\frac{d}{d+1}} \Big) \Big(1 -d\big(\frac{K}{n}\big)^{\frac{1}{d+1}} \Big) \Big]\\
&\geq \max\Big[\bar S , \lambda_{K,k} \Big(n - (2 + 2\frac{A}{\Sigma_K} +d) K^{\frac{1}{d+1}} n^{\frac{d}{d+1}} \Big) \Big].
\end{align*}

We call $N =n - (2 + 2\frac{A}{\Sigma_K} +d) K^{\frac{1}{d+1}} n^{\frac{d}{d+1}} $ in the sequel. Note that $\forall k$, we have $S_{k} \geq \max[\bar S,\lambda_{K,k} N]$.

Note also that for $\delta\leq 1$, we have 
\begin{align*}
A &= (2L\sqrt{d}\sqrt{\log(2K/\delta)})^{\frac{d}{d+1}}\\
&\leq 4 (L+1) \sqrt{d} \sqrt{\log(K/\delta)}.
\end{align*}
We thus have that 

\begin{align}\label{eq:defN}
n \geq N \geq n - 7(L+1)d^{3/2} \sqrt{\log(K/\delta)} (1+\frac{1}{\Sigma_K}) K^{\frac{1}{d+1}} n^{\frac{d}{d+1}}.
\end{align}

\section{Proof of Theorem~\ref{th:asymp.conv}}\label{proof:asymp.conv}

\paragraph{Step 1: Notations}

Let $\big((\Omega^n_{k})_{k \leq K_n}\big)_n$ be a sequence of partitions in hyper-cubic strata of same measure. Let us also assume that the number of strata $K_n$ in partition $(\Omega^n_{k})_k$ is such that $\lim_{n \rightarrow +\infty} K_n = +\infty$ and $\lim_{n \rightarrow \infty} \frac{K_n^{d+2} \log(n)^{d+3}}{n^{d+1}} = 0$. On each of those partitions, $MC-UCB$ is launched with respectively $n$ samples and parameter $\delta_n = \frac{1}{n^2}$.

The number of hyper-cubic sub-strata built by the algorithm in stratum $\Omega^n_{k}$ is $S_{n,k}$. Let us write $\Big(\big((\Omega^n_{k,s})_{s\leq S_{n,k}}\big)_{k \leq K_n}\Big)_n$ the partition in hyper-cubic strata formed with those sub-strata. By construction of the algorithm, there is at least one point per sub-stratum. The estimate of the mean of the function is built with the first point in each of those sub-strata.

Let us write $g^{(1)}_n(x) =  \sum_{k=1}^{K_n} \sum_{s=1}^{S_{n,k}} \frac{\si_{n,k,s}^2}{w_{n,k,s}^{1/2d}} \ind{\Omega^n_{k,s}}(x)=  \sum_{k=1}^{K_n} \sum_{s=1}^{S_{n,k}} \si_{n,k,s}^2 \frac{S_{n,k}^{1/2d}}{w_{n,k}^{1/2d}} \ind{\Omega^n_{k,s}}(x)$. From step 1 of the proof of Lemma~\ref{prop:asymbound}, it converges with $n$ (because $K_n \rightarrow +\infty$ when $n  \rightarrow \infty$ and thus the diameter of each stratum goes to $0$)  point-wise to $\frac{||\nabla f(x)||_2^2}{12}$.

Let us write $g^{(2)}_n(x) =  \sum_{k=1}^{K_n} \frac{\si_{n,k}^2}{w_{n,k}^{1/2d}} \ind{\Omega^n_{k}}(x)$. From step 1 of the proof of Lemma~\ref{prop:asymbound}, it converges with $n$  point-wise to $\frac{||\nabla f(x)||_2^2}{12}$. This convergence implies, as $||\nabla f||_2^2$ is bounded and thus as $\int ||\nabla f||_2^{\frac{d}{d+1}}$ is bounded, by the Theorem of Dominated convergence that $\lim_{n \rightarrow + \infty} \Sigma_{K_n}= \lim_{n \rightarrow + \infty} \int_{[0,1]^d} (g_n^{(2)}(x))^{\frac{d}{2(d+1)}} dx = \int_{[0,1]^d} (\frac{||\nabla f(x)||_2}{12})^{\frac{d}{(d+1)}} dx >0$.

Define $\lambda_n(x) = \sum_{k=1}^{K_n} \frac {\lambda_{K_n,k}}{w_{n,k}} \ind{\Omega^n_{k}} = \sum_{k=1}^{K_n} \frac{(w_{n,k} \si_{n,k})^{\frac{d}{d+1}}}{w_{n,k} \Sigma_{K_n}} \ind{\Omega^n_{k}}= \frac{(g_n(x))^{\frac{d}{2(d+1)}} }{ \Sigma_{K_n}}$. We thus know, as the limit of $(\Sigma_{K_n})_n$ exists and is bigger than $0$, that $\lambda_n(x)$ converges pointwise to $s(x) = \frac{||\nabla f(x)||_2^{\frac{d}{d+1}}}{\int_{[0,1]^d} ||\nabla f(x)||_2^{\frac{d}{(d+1)}} dx}$.

Let us also define $s_{n}(x) = \sum_{k=1}^{K_n} \frac{S_{n,k}}{nw_{n,k}} \ind{\Omega^n_{k}}(x)$.

-\paragraph{Step 1: Majoration of of $\frac{1}{s_n}$.}

Let us consider only functions $f$ that are not everywhere constant on the domain, as otherwise the bound on the pseudo-risk is trivial\footnote{If the function is everywhere constant, the samples are always equal to the integral, and the pseudo-risk of the estimate is zero.}. Then $\exists \X \in [0,1]^d$ such that $\X$ is measurable and such that $\int_{\X} 1 >0$, and such that $ \forall x \in \X, ||\nabla f(x)||_2 >0$. Then $\int_{[0,1]^d} (\frac{||\nabla f(x)||_2}{12})^{\frac{d}{(d+1)}} dx >0$.
\\
Let $N_n$ be defined as in the proof of Lemma~\ref{lem:nbpulls}, i.e.~$N_n$ as in Equation~\ref{eq:defN}.
As $\lim_{n \rightarrow + \infty} \Sigma_{K_n} = \int_{[0,1]^d} (\frac{||\nabla f(x)||_2}{12})^{\frac{d}{(d+1)}} dx$, we know that for any $n$ sufficiently large, $\lim_n \Sigma_{K_n} \geq  \frac{1}{2} \int_{[0,1]^d} (\frac{||\nabla f(x)||_2}{12})^{\frac{d}{(d+1)}} dx$. We thus have
\begin{align*}
n \geq N_n &\geq n - 7(L+1)d^{3/2} \sqrt{\log(K_n/\delta_n)} (1+\frac{1}{\Sigma_{K_n}}) K^{\frac{1}{d+1}} n^{\frac{d}{d+1}}\\
&\geq n - C \sqrt{\log(K_n n^2)} K_n^{\frac{1}{d+1}} n^{\frac{d}{d+1}},
\end{align*}
with $C <+\infty$ as $\int_{[0,1]^d} (\frac{||\nabla f(x)||_2}{12})^{\frac{d}{(d+1)}} dx >0$. As by definition of the sequence of partitions, $\lim_{n \rightarrow +\infty}\sqrt{\log(K_n n^2)} \Big(\frac{K_n}{n}\big)^{\frac{1}{d+1}} =0$, we know that $\lim_{n \rightarrow +\infty} \frac{N_n}{n} = 1$.

By Lemma~\ref{lem:nbpulls}, with probability $1-\delta_n$, $\forall k, S_{n,k} \geq \lambda_{K_n, k} N_n$. We thus have
\begin{align*}
 \P\left(  \frac{1}{s_n(x)} - \frac{1}{\lambda_n(x)} \geq \frac{1}{ \lambda_n(x)} (\frac{n}{N_n} -1)  \right) \leq \delta_n,
\end{align*}
which leads to
\begin{align*}
 \P\left(  \frac{1}{s_n(x)}  \geq \frac{1}{ \lambda_n(x)} \frac{n}{N_n}  \right) \leq \delta_n.
\end{align*}

Let $\X^+ =\{x \in [0,1]^d:  ||\nabla f||_2>0 \}$. By the last Equation, $\forall \epsilon>0$, $\forall x \in \X^+$, for n sufficiently large ($\exists n'$ such that $\forall n  \geq n'$), $\P(\frac{1}{s_n(x)} - \frac{1}{\lambda_n(x)} \geq \epsilon) \leq \delta_n $. Note that $\sum_{n=1}^{+ \infty} \delta_n = \sum_{n=1}^{+ \infty} \frac{1}{n^2} \leq + \infty$. We can thus use Borel-Cantelli's Theorem and this gives us that on $\X^+$, $\lim \sup_n \frac{1}{s_n(x)} - \frac{1}{\lambda_n(x)} \leq 0$ a.s..

We thus deduce (i) by the definition of $\lambda_n$ and the fact that it converges almost surely to $s$ and (ii) by the fact that $\lim_n \frac{N_n}{n}=1$, that $\lim \sup_n \frac{1}{ \lambda_n(x)} \leq \frac{1}{s(x)}$ a.s. (since, by definition, $s_n(x) \geq \frac{\bar S}{n w_{n,K}} > 0$).

From that we deduce that $\forall x \in \X^+$, $\lim \sup_n \frac{1}{s_n(x)} \leq \frac{1}{s(x)}$ a.s.. As on $[0,1]^d - \X^+$, $s(x) = 0$, we have $\forall x \in [0,1]^d$, that $\lim \sup_n \frac{1}{s_n(x)} \leq \frac{1}{s(x)}$ a.s..

\paragraph{Step 2: Convergence rate of the pseudo-risk.}


%
%
%
%
The pseudo-risk of the estimate $\hat \mu_n$ is
\begin{align*}
 \sum_{k=1}^{K_n} \sum_{s=1}^{S_{n,k}} \Big(\frac{w_{n,k}}{S_{n,k}}\Big)^2 \si_{n,k,s}^2 &= n^{1+2/d}\int_{[0,1]^d} g^{(1)}_{n}(x) \frac{1}{s_{n}(x)^{1+2/d}} dx.
\end{align*}

On $[0,1]^d$, $g^{(1)}_n$ converges pointwise to $\frac{||\nabla f||_2^2}{12}$, and $\lim \sup_{n \rightarrow +\infty}\frac{1}{s_{n}(x)^{1+2/d}} \leq \frac{1}{s(x)^{1+2/d}}$ a.s. We finally have by Fatou's Lemma that
\begin{align*}
 \int_{[0,1]^d} g^{(1)}_{n}(x) \frac{1}{s_{n}(x)^{1+2/d}} dx &\leq \int_{[0,1]^d} \lim \sup_n \Big(g^{(1)}_{n}(x) \frac{1}{s_{n}(x)^{1+2/d}}\Big) dx\\
&\leq \int_{[0,1]^d} \lim \sup_n g^{(1)}_{n}(x) \lim \sup_n\frac{1}{s_{n}(x)^{1+2/d}} dx\\
&\leq \int_{[0,1]^d} \frac{||\nabla f||_2^2}{12} \frac{1}{s(x)^{1+2/d}} dx.
\end{align*}

By plugging in the last Equation the Definition of $s$, we conclude the proof.

\section{Proof of Theorems~\ref{th:pi}}\label{proof:thpi}

%

\paragraph{Step 0: Some inequalities when the second derivative of $f$ is bounded}
Let $a$ be a point in $\Omega$.

$f$ admits a Taylor expansion in any point.
For any $x\in \Omega$ have $|f(x)- f(a) + \nabla f(a).(x-a)| \leq M ||x-a||_2^2$ with $2M$ a bound of the second derivative of $f$.
\\
Note also that $||\nabla f(x) - \nabla f(a)||_2 \leq M||x-a||_2$.


Note also that
\begin{align*}
 \Big| ||\nabla f(x)||_2^2 - ||\nabla f(a)||_2^2 \Big| &\leq  \Big| \big(||\nabla f(x)||_2\big)^2 - ||\nabla f(a)||_2^2 \Big|\\
&\leq \Big| \big(||\nabla f(a)||_2 + M ||x-a||_2\big)^2 - ||\nabla f(a)||_2^2 \Big|\\
&\leq \Big| ||\nabla f(a)||_2^2 + 2M||\nabla f(a)||_2 ||x-a||_2 + M^2||x-a||_2^2 - ||\nabla f(a)||_2^2 \Big|\\
&\leq 2M||\nabla f(a)||_2 ||x-a||_2 + M^2||x-a||_2^2.
\end{align*}
This means that
\begin{align}\label{eq:distvar}
\Big| ||\nabla f(x)||_2 - ||\nabla f(a)||_2 \Big| \leq M||x-a||_2.
\end{align}
\\
\paragraph{Step 1: Variance on a small box}
Let us place us on one small box of size $w$ and such that the corresponding domain is $\Omega_w = \prod [a_i-\frac{w^{1/d}}{2}, a_i+\frac{w^{1/d}}{2}]$. We can do a Taylor expansion in $a$ and have
\begin{align*}
 |f(x) - f(a) + \nabla f(a) (x -a)| \leq M ||x-a||_2^2,
\end{align*}
with $2M$ a bound of the second derivative of $f$.

Note that because of the previous equation
\begin{align}
 |\frac{1}{w} \int_{\Omega_w} \Big(f(u)  - f(a) +  \nabla f(a) (u -a)\Big)du| &\leq \frac{1}{w} \int_{\Omega_w} |f(u) - f(a) + \nabla f(a) (u -a)| du \nonumber\\
&\leq M ||x-a||_2^2. \label{eq:1}
\end{align}
This implies because $a_i = \int_{a_i-\frac{w^{1/d}}{2}}^{a_i+\frac{w^{1/d}}{2}} u du$ that
\begin{align}
 |\frac{1}{w} \int_{\Omega_w}  f(u) du  - f(a) | \leq M ||x-a||_2^2. \label{eq:2}
\end{align}
Finally, by combining Equations~\ref{eq:1} and~\ref{eq:2}, we get
\begin{align*}
 |f(x) - \frac{1}{w} \int_{\Omega_w}  f(u) du  + \nabla f(a) (x -a)| \leq 2M ||x-a||_2^2.
\end{align*}
Triangle inequality on the last Equation leads to
\begin{align*}
 |f(x) - \frac{1}{w} \int_{\Omega_w}  f(u) du|  \leq |\nabla f(a) (x -a)| + 2M ||x-a||_2^2.
\end{align*}
This means by integrating that
\begin{align}
 \int_{\Omega_w} \Big(f(x) - \frac{1}{w} \int_{\Omega_w}  f(u) du \Big)^2dx  \leq& \int_{\Omega_w} \Big(|\nabla f(a) (x -a)| + 2M ||x-a||_2^2 \Big)^2 dx \nonumber\\
\leq& \int_{\Omega_w} \Big(\nabla f(a) (x -a)\Big)^2 dx \label{eq:term1}\\
&+ 2M\int_{\Omega_w} \Big(\nabla f(a) (x -a)|\Big)||x-a||_2^2 dx \label{eq:term2}\\
&+ 4M^2 \int_{\Omega_w} ||x-a||_2^4 dx. \label{eq:term3}
\end{align}

Note first that because $a_i = \int_{a_i-\frac{w^{1/d}}{2}}^{a_i+\frac{w^{1/d}}{2}} u du$, we have for the term in Equation~\ref{eq:term1}
\begin{align}
\int_{\Omega_w} \Big(\nabla f(a) (x -a)\Big)^2 dx &= \int_{\Omega_w} \Big(\sum_{i=1}^d \nabla f(a)_i (x_i -a_i)\Big)^2 dx \nonumber\\
&=  w^{1-1/d}\sum_{i=1}^d \int_{a_i - \frac{w^{1/d}}{2}}^{a_i+\frac{w^{1/d}}{2}} \nabla f(a)_i^2 (x_i -a_i)^2 dx_i \nonumber\\
&=  \sum_{i=1}^d \nabla f(a)_i^2 \frac{w^{1+2/d}}{12} \nonumber\\
& = \frac{w^{1+2/d}}{12} ||\nabla f(a)||_2^2. \label{eq:term11}
\end{align}

Now note that for the term in Equation~\ref{eq:term3}
\begin{align}
 \int_{\Omega_w} ||x-a||_2^4 dx &= \int_{\Omega_w} \Big(\sum_{i=1}^d (x_i - a_i)^2\Big)^2 dx \nonumber\\
&\leq d^2 w^{1+4/d}. \label{eq:term33}
\end{align}

Now note that because of Cauchy-Schwartz and by using Equations~\ref{eq:term11} and~\ref{eq:term33}, we have for the term in Equation~\ref{eq:term2}
\begin{align}
 \int_{\Omega_w} \Big(\nabla f(a) (x -a)|\Big)||x-a||_2^2 dx &\leq \sqrt{\int_{\Omega_w} \Big(\nabla f(a) (x -a)|\Big)^2 dx}\sqrt{\int_{\Omega_w}||x-a||_2^4 dx} \nonumber\\
&\leq ||\nabla f(a)||_2 w^{1/2 + 1/d} \sqrt{d^2w^{1+4/d}} \nonumber\\
&\leq d||\nabla f(a)||_2 w^{1 + 3/d}. \label{eq:term22}
\end{align}

We thus have by combining Equations~\ref{eq:term1}, \ref{eq:term2}, \ref{eq:term3}, \ref{eq:term11}, \ref{eq:term22} and~\ref{eq:term33}
\begin{align*}
 \int_{\Omega_w} \Big(f(x) - \frac{1}{w} \int_{\Omega_w}  f(u) du \Big)^2dx
&\leq \frac{||\nabla f(a)||_2^2}{12}w^{1+2/d}  + 2M d ||\nabla f(a)||_2 w^{1 + 3/d} + 4M^2d^2 w^{1+4/d}.
\end{align*}
This leads to using Step 0 in Proof~\ref{proof:asymbound}
\begin{align}
 w^2 \si^2  &\leq \frac{||\nabla f(a)||_2^2}{12}w^{2+2/d}  + 2M d ||\nabla f(a)||_2 w^{2 + 3/d} + 4M^2 d^2 w^{2+4/d} \nonumber\\
&= w^{2+2/d} \big(\frac{||\nabla f(a)||_2}{2\sqrt{3}} + 2M d w^{1/d}  \big)^2. \label{eq:majorstrat}
\end{align}

In the same way, one can prove
\begin{align}
 w^2 \si^2  \geq  w^{2+2/d} \big(\frac{||\nabla f(a)||_2}{2\sqrt{3}} - 2M d w^{1/d}  \big)^2. \label{eq:minorstrat}
\end{align}

\paragraph{Step 2: Majoration on the strata}

Lemma~\ref{lem:nbpulls} tells us that with probability $1-\delta$ (i.e.~on the event $\xi$), each stratum $\Omega_k$ is partitioned in $S_{k} \geq \max\Bigg[\lambda_{p,K}N, \bar S\Bigg]$ hyper-cubic substrata $\Omega_{k,i}$ of same measure, and that that there is at least one sample per stratum.
The measure of those sub-strata is thus $w_{k,i} = \frac{w_k}{S_{k}}$.

We have for stratum $\Omega_{k,i}$ by using Equation~\ref{eq:majorstrat}
\begin{align*}
 w_{k,i}^2 \si_{k,i}^2 &\leq w_{k,i}^{2+2/d} \big(\frac{||\nabla f(a_{k,i})||_2}{2\sqrt{3}} + 2M d w_{k,i}^{1/d}  \big)^2,
\end{align*}
where $a_{k,i}$ is the center of stratum $\Omega_{k,i}$.

Let $c_{k,i}$ be a point in $\Omega_{k,i}$ such that $c_{k,i} = \arg \min_{c \in \Omega_{k,i}} ||\nabla f (c)||_2$. 
By using that and Equation~\ref{eq:distvar}, we get that the variance on strata $k$ that is bounded by
%
%
\begin{align*}
 \sum_{i=1}^{S_{k}} w_{k,i}^2 \si_{k,i}^2 &\leq \sum_{i=1}^{S_{k}} w_{k,i}^{2+2/d} \big(\frac{||\nabla f(a_{k,i})||_2}{2\sqrt{3}} + 2M d w_{k,i}^{1/d}  \big)^2\\
\leq& \sum_{i=1}^{S_{k}} w_{k,i}^{2+2/d} \big(\frac{||\nabla f(c_{k,i})||_2}{2\sqrt{3}} +  3M d w_{k,i}^{1/d}  \big)^2\\
\leq& \frac{w_k}{S_{k}} \sum_{i=1}^{S_{k}}  w_{k,i}^{\frac{d+2}{d}} \big(\frac{||\nabla f(c_{k,i})||_2}{2\sqrt{3}} +  3M d w_{k,i}^{1/d}  \big)^2.
\end{align*}

Let us call $g(x) = \frac{||\nabla f(x)||_2}{2\sqrt{3}} +  3M d w_{k}^{1/d}$. As $w_k \geq w_{k,i}$, and $||\nabla f||_2$ is positive, we have
\begin{align}
 \sum_{i=1}^{S_{k}} w_{k,i}^2 \si_{k,i}^2 
\leq& \frac{w_k}{S_{k}} \sum_{i=1}^{S_{k}}  w_{k,i}^{\frac{d+2}{d}} g(c_{k,i})^2. \label{eq:varstratk}
\end{align}

\paragraph{Step 3: Minoration of the number of sub-strata in each stratum}

By setting Equation~\ref{eq:majorstrat} to the power $\frac{d}{2(d+1)}$, we get on stratum $\Omega_k$ that
\begin{align*}
 (w_k \sigma_k)^{\frac{d}{d+1}}
&\leq w_k \big(\frac{||\nabla f(a_k)||_2}{2\sqrt{3}} + 2M d w_k^{1/d}\big)^{\frac{d}{d+1}}.
\end{align*}

Let $c_{k}^m$ be a point in $\Omega_{k}$ such that $c_{k}^m = \arg \min_{c \in \Omega_{k}} ||\nabla f(c)||_2$. Note that this implies that $\sum_{k=1}^{K} w_k \big(\frac{||\nabla f(c_{k}^m)||_2}{2\sqrt{3}} + 3M d w_{k}^{1/d}  \big)^{\frac{d}{d+1}} \leq \int_{[0,1]^d} \big(\frac{||\nabla f(u)||_2}{2\sqrt{3}} + 3M d w_{k}^{1/d}  \big)^{\frac{d}{d+1}} du$. By using that and Equation~\ref{eq:distvar}, we get that $\Sigma_K = \sum_k  (w_k \sigma_k)^{\frac{d}{d+1}}$ is bounded as
%
%
\begin{align}
\Sigma_K &\leq \sum_{k=1}^{K} w_k \big(\frac{||\nabla f(a_k)||_2}{2\sqrt{3}} + 2M d w_k^{1/d}\big)^{\frac{d}{d+1}} \nonumber\\
\leq& \sum_{k=1}^{K} w_k \big(\frac{||\nabla f(c_k^m)||_2}{2\sqrt{3}} + 3M d w_k^{1/d}\big)^{\frac{d}{d+1}} \nonumber\\
 \leq& \int_{[0,1]^d} \big(\frac{||\nabla f(u)||_2}{2\sqrt{3}} + 3M d w_{k}^{1/d}  \big)^{\frac{d}{d+1}} du \nonumber\\
\leq& \int_{[0,1]^d} g(u)^{\frac{d}{d+1}} du. \label{eq:SiK}
\end{align}

In the same way, we can deduce
\begin{align}
\Sigma_K  \geq& \int_{[0,1]^d} \big(\frac{||\nabla f(u)||_2}{2\sqrt{3}} - 3M d w_{k}^{1/d}  \big)^{\frac{d}{d+1}} du. \label{eq:SiK2}
\end{align}

Let $c_{k}^M$ be a point in $\Omega_{k}$ such that $c_{k}^M = \arg \max_{c \in \Omega_{k}} ||\nabla f(c)||_2$. 
For a stratum $k$, by using Equations~\ref{eq:minorstrat} and~\ref{eq:distvar}
\begin{align*}
(w_k \si_k)^{\frac{d+2}{d+1}}  &\geq w_k^{\frac{d+2}{d}} \big(\frac{||\nabla f(a_k)||_2}{2\sqrt{3}} - 2M d w_k^{1/d}\big)^{\frac{d+2}{d+1}}\\
&\geq w_k^{\frac{d+2}{d}} \big(\frac{||\nabla f(c_k^M)||_2}{2\sqrt{3}} - 3M d w_k^{1/d}\big)^{\frac{d+2}{d+1}}.
\end{align*}
As for any $u>0$ and $\alpha>0$ one has $(1-u)^{-\alpha} \geq 1 + \alpha u$, the last Equation leads to
\begin{align*}
\frac{1}{(w_k \si_k)^{\frac{d+2}{d+1}}}  &\leq \frac{1}{w_k^{\frac{d+2}{d}} \big(\frac{||\nabla f(c_k^M)||_2}{2\sqrt{3}} + 3M d w_{k}^{1/d} - 3M d (w_k^{1/d} + w_{k}^{1/d}) \big)^{\frac{d+2}{d+1}}}\\
&\leq \frac{1}{w_k^{\frac{d+2}{d}} \big(g(c_k^M) - 6M d w_k^{1/d} \big)^{\frac{d+2}{d+1}}}\\
&\leq \frac{1}{w_k^{\frac{d+2}{d}} g(c_k^M)^{\frac{d}{d+1}} \big(1 - \frac{6M d w_k^{1/d}}{g(c_k^M)}\big)^{\frac{d+2}{d+1}}}\\
&\leq \frac{1}{w_k^{\frac{d+2}{d}} \big(g(c_k^M)\big)^{\frac{d+2}{d+1}}} \big(1 + {(\frac{d+2}{d+1}}) \frac{6M d w_k^{1/d}}{g(c_k^M)}\big)\\
&\leq \frac{1}{w_k^{\frac{d+2}{d}} } \big(\frac{1}{\big(g(c_k^M)\big)^{\frac{d+2}{d+1}}} +  \frac{9 M d w_k^{1/d}}{(g(c_k^M))^{\frac{2d+3}{d+1}}}\big).
\end{align*}

As $w_{k,i} = \frac{w_k}{S_k}$ this leads with the last Equation and Equation~\ref{eq:SiK}
\begin{align}\label{eq:wki}
(w_{k,i})^{\frac{d+2}{d}}&\leq \Big(\frac{\int_{[0,1]^d} \big(g(u) \big)^{\frac{d}{d+1}} du}{N} \Big)^{\frac{d+2}{d}} \big(\frac{1}{\big(g(c_k^M)\big)^{\frac{d+2}{d+1}}} +  \frac{9 M d w_k^{1/d}}{(g(c_k^M))^{\frac{2d+3}{d+1}}}\big).
\end{align}


\paragraph{Step 4: Bound on the pseudo-risk}

As $c_k^M = \max_{c \in  \Omega_k} ||\nabla f(c)||_2$ and $c_{k,i} = \min_{c \in  \Omega_{k,i}} ||\nabla f(c)||_2$, and as $g(x) = \frac{||\nabla f(x)||_2}{2\sqrt{3}} +  3M d w_{k}^{1/d}$, we have for any $(a,b) \geq 0$ that $\frac{g(c_{k,i})^a}{g(c_{k}^M)^b} \leq \min_{c \in \Omega_{k,i}} g(c)^{a-b}$. By using that and Equations~\ref{eq:varstratk} and~\ref{eq:wki}
\begin{align*}
 \sum_{i=1}^{S_{k}} w_{k,i}^2 \si_{k,i}^2  \leq& \frac{w_k}{S_{k}} \Big(\frac{\int_{[0,1]^d} \big(g(u) \big)^{\frac{d}{d+1}} du}{N} \Big)^{\frac{d+2}{d}}  \sum_{i=1}^{S_{k}}  w_{k,i}^{\frac{d+2}{d}} g(c_{k,i})^2\\
\leq& \Big(\frac{\int_{[0,1]^d} \big(g(u) \big)^{\frac{d}{d+1}} du}{N} \Big)^{\frac{d+2}{d}} \frac{w_k}{S_{k}} \sum_{i=1}^{S_{k}}    \big(\frac{1}{\big(g(c_k^M)\big)^{\frac{d+2}{d+1}}} +  \frac{9 M d w_k^{1/d}}{(g(c_k^M))^{\frac{2d+3}{d+1}}}\big) g(c_{k,i})^2\\
\leq& \Big(\frac{\int_{[0,1]^d} \big(g(u) \big)^{\frac{d}{d+1}} du}{N} \Big)^{\frac{d+2}{d}} \frac{w_k}{S_{k}} \sum_{i=1}^{S_{k}}    \big( \min_{c \in \Omega_{k,i}} g(c)^{\frac{d}{d+1}} +  \min_{c \in \Omega_{k,i}} \frac{9 M d w_k^{1/d}}{(g(c))^{\frac{1}{d+1}}}\big).
\end{align*}

Note also that by definition, $g(x) \geq 3Mdw_k^{1/d}$. From that and the previous Equation, we deduce
\begin{align*}
 \sum_{i=1}^{S_{k}} w_{k,i}^2 \si_{k,i}^2 \leq& \Big(\frac{\int_{[0,1]^d} \big(g(u) \big)^{\frac{d}{d+1}} du}{N} \Big)^{\frac{d+2}{d}} \frac{w_k}{S_{k}} \sum_{i=1}^{S_{k}}    \big( \min_{c \in \Omega_{k,i}} g(c)^{\frac{d}{d+1}} +   \frac{9 M d w_k^{1/d}}{(3Mdw_k^{1/d})^{\frac{1}{d+1}}}\big)\\
\leq& \Big(\frac{\int_{[0,1]^d} \big(g(u) \big)^{\frac{d}{d+1}} du}{ N} \Big)^{\frac{d+2}{d}} w_k  \big( \frac{1}{w_k}\int_{\Omega_k} g(u)^{\frac{d}{d+1}}du +   9 M d w_k^{\frac{1}{d+1}}\big).
\end{align*}

Finally, by summing over all strata and because all strata have same measure $w_k = \frac{1}{K}$
\begin{align}
 \sum_{i=1}^K \sum_{i=1}^{S_{k}} w_{k,i}^2 \si_{k,i}^2 \leq& \Big(\frac{\int_{[0,1]^d} \big(g(u) \big)^{\frac{d}{d+1}} du}{N} \Big)^{\frac{d+2}{d}}  \sum_{k=1}^K\big( \int_{\Omega_k} g(u)^{\frac{d}{d+1}}du +  w_k \times 9 M d w_k^{\frac{1}{d+1}}\big) \nonumber\\
\leq& \Big(\frac{\int_{[0,1]^d} \big(g(u) \big)^{\frac{d}{d+1}} du}{N} \Big)^{\frac{d+2}{d}}  \big( \int_{[0,1]^d} g(u)^{\frac{d}{d+1}}du +  9 M d \big(\frac{1}{K}\big)^{\frac{1}{d+1}}\big) \nonumber \\
\leq& \frac{1}{N^{\frac{d+2}{d}}}  \Big( \big(\int_{[0,1]^d} g(u)^{\frac{d}{d+1}}du\big)^{\frac{2(d+1)}{d}} +  9 M d \big(\int_{[0,1]^d} g(u)^{\frac{d}{d+1}}du\big)^{\frac{d+2}{d}} \big(\frac{1}{K}\big)^{\frac{1}{d+1}}\Big) \label{eq:risk1}.
\end{align}

\paragraph{Step 5: Bound on $\int_{[0,1]^d} g(u)^{\frac{d}{d+1}}du$}

Note that because $\frac{d}{d+1} \leq 1$, we have
\begin{align*}
g(u)^{\frac{d}{d+1}} &= \Big(\frac{||\nabla f(u)||_2}{2\sqrt{3}} +  3M d w_{k}^{1/d}\Big)^{\frac{d}{d+1}}\\
&\leq \big(\frac{||\nabla f(u)||_2}{2\sqrt{3}}\big)^{\frac{d}{d+1}} +  3M d w_{k}^{\frac{1}{d+1}}
\end{align*}

We thus have
\begin{align}\label{eq:bg}
\int_{[0,1]^d} g(u)^{\frac{d}{d+1}} du &\leq \int_{[0,1]^d} \big(\frac{||\nabla f(u)||_2}{2\sqrt{3}}\big)^{\frac{d}{d+1}} du +  3M d w_{k}^{\frac{1}{d+1}}.
\end{align}

Note also that for $x \geq 0$, and as $\frac{2(d+1)}{d} \leq 4$, we have
\begin{align*}
 (1+x)^{\frac{2(d+1)}{d}} \leq&  (1+x)^{4} \leq  1 + 2^4 \max (x, x^2, x^3, x^4).
\end{align*}
Let us call $\Sigma = \int_{[0,1]^d} \big(\frac{||\nabla f(u)||_2}{2\sqrt{3}}\big)^{\frac{d}{d+1}} du$. Then by applying the previous result to Equation~\ref{eq:bg}, we get
\begin{align}
\Big(\int_{[0,1]^d} g(u)^{\frac{d}{d+1}} du\Big)^{\frac{2(d+1)}{d}} &\leq \Big(\int_{[0,1]^d} \big(\frac{||\nabla f(u)||_2}{2\sqrt{3}}\big)^{\frac{d}{d+1}} du +  3M d w_{k}^{\frac{1}{d+1}}\Big)^{\frac{2(d+1)}{d}} \nonumber\\
&= \Sigma^{\frac{2(d+1)}{d}}\Big(1 +  \frac{3M d}{\Sigma} w_{k}^{\frac{1}{d+1}}\Big)^{\frac{2(d+1)}{d}} \nonumber\\
&\leq \Sigma^{\frac{2(d+1)}{d}} + 16 \Sigma^{\frac{2(d+1)}{d}}\Big(1 +  \frac{3M d}{\Sigma} \Big)^4 w_{k}^{\frac{1}{d+1}}. \label{eq:bg3}
\end{align}

Note also that by Equation~\ref{eq:distvar}, we know that $||\nabla f(u)||_2 \leq ||\nabla f(0)||_2 + M\sqrt{d} $. From that we deduce that
\begin{align}
\int_{[0,1]^d} g(u)^{\frac{d}{d+1}} du &\leq \Sigma +  3M d w_{k}^{\frac{1}{d+1}}\nonumber\\
&\leq \Sigma+ 3Md. \label{eq:bg2}
\end{align}

\paragraph{Step 6: Final bound on the pseudo-risk}

From Equations~\ref{eq:risk1}, \ref{eq:bg3} and~\ref{eq:bg2}, we deduce
\begin{align*}
 \sum_{i=1}^K \sum_{i=1}^{S_{k}} w_{k,i}^2 \si_{k,i}^2 
\leq& \frac{1}{N^{\frac{d+2}{d}}}  \Big( \big(\int_{[0,1]^d} g(u)^{\frac{d}{d+1}}du\big)^{\frac{2(d+1)}{d}} +  9 M d \big(\int_{[0,1]^d} g(u)^{\frac{d}{d+1}}du\big)^{\frac{d+2}{d}} \big(\frac{1}{K}\big)^{\frac{1}{d+1}}\Big)\\
\leq&  \frac{1}{N^{\frac{d+2}{d}}}  \Big[ \Sigma^{\frac{2(d+1)}{d}} + 16 \Sigma^{\frac{2(d+1)}{d}}\Big(1 +  \frac{3M d}{\Sigma} \Big)^4 w_{k}^{\frac{1}{d+1}}\\
 &+  9 M d \big( \Sigma + 3Md \big)^{\frac{d+2}{d}} \big(\frac{1}{K}\big)^{\frac{1}{d+1}}\Big]\\
\leq&  \frac{1}{N^{\frac{d+2}{d}}}  \Big[ \Sigma^{\frac{2(d+1)}{d}} + 25 Md (\Sigma+1)^{\frac{2(d+1)}{d}}\Big(1 +  \frac{3M d}{\Sigma} \Big)^4 \big(\frac{1}{K}\big)^{\frac{1}{d+1}}\Big]\\
\leq&  \frac{1}{N^{\frac{d+2}{d}}}  \Big[ \Sigma^{\frac{2(d+1)}{d}} + C \big(\frac{1}{K}\big)^{\frac{1}{d+1}}\Big],
\end{align*}
where $C = 25 Md (\Sigma+1)^{\frac{2(d+1)}{d}}\Big(1 +  \frac{3M d}{\Sigma} \Big)^4$.

Note that $N =  n - (2 + 2\frac{A}{\Sigma_K} +d) K^{\frac{1}{d+1}} n^{\frac{d}{d+1}} = n - B K^{\frac{1}{d+1}} n^{\frac{d}{d+1}}$, where $B=2 + 2\frac{A}{\Sigma_K} +d$. From plugging that in the last Equation, we get
\begin{align*}
 \sum_{i=1}^K \sum_{i=1}^{S_{k}} w_{k,i}^2 \si_{k,i}^2 
\leq&  \frac{1}{\Big(n - B K^{\frac{1}{d+1}} n^{\frac{d}{d+1}}\Big)^{\frac{d+2}{d}}}  \Big[ \Sigma^{\frac{2(d+1)}{d}} + C \big(\frac{1}{K}\big)^{\frac{1}{d+1}}\Big]\\
\leq&  \frac{1}{n^{\frac{d+2}{d}}\Big(1 - B K^{\frac{1}{d+1}} n^{-\frac{1}{d+1}}\Big)^{\frac{d+2}{d}}}  \Big[ \Sigma^{\frac{2(d+1)}{d}} + C \big(\frac{1}{K}\big)^{\frac{1}{d+1}}\Big]\\
\leq&  \frac{1}{n^{\frac{d+2}{d}}} \Big[ 1 +  (\frac{d+2}{d} )B K^{\frac{1}{d+1}} n^{-\frac{1}{d+1}}\Big]  \Big[ \Sigma^{\frac{2(d+1)}{d}} + C \big(\frac{1}{K}\big)^{\frac{1}{d+1}}\Big]\\
\leq&  \frac{1}{n^{\frac{d+2}{d}}} \Big[ \Sigma^{\frac{2(d+1)}{d}} +  3\Sigma^{\frac{2(d+1)}{d}} B K^{\frac{1}{d+1}} n^{-\frac{1}{d+1}} + C \big(\frac{1}{K}\big)^{\frac{1}{d+1}} + 3 B C n^{-\frac{1}{d+1}} \Big],
\end{align*}
where we use for passing from the second to the third line of the Equation that $(1-u)^{-\alpha} \leq 1+\alpha u$.

By it's definition, $C \geq \Sigma^{\frac{2(d+1)}{d}}$ and this leads to
\begin{align}\label{eq:alfin}
 \sum_{i=1}^K \sum_{i=1}^{S_{k}} w_{k,i}^2 \si_{k,i}^2 
\leq&  \frac{1}{n^{\frac{d+2}{d}}} \Big[ \Sigma^{\frac{2(d+1)}{d}} +  6BC K^{\frac{1}{d+1}} n^{-\frac{1}{d+1}} + C \big(\frac{1}{K}\big)^{\frac{1}{d+1}} \Big].
\end{align}


Note first that by Equation~\ref{eq:SiK2} and because $||\nabla f||_2 \leq L$ we have
\begin{align*}
\Sigma_K  \geq& \int_{[0,1]^d} \big(\frac{||\nabla f(u)||_2}{2\sqrt{3}} - 3M d w_{k}^{1/d}  \big)^{\frac{d}{d+1}} du\\
\geq& \Sigma - 3LM d w_{k}^{\frac{1}{d+1}}.
\end{align*}
From that we deduce that 
\begin{align*}
B &\leq 2 + 2\frac{4 (L+1) \sqrt{d} \sqrt{\log(K/\delta)}}{\Sigma - 3LM d w_{k}^{\frac{1}{d+1}}} +d \\
&\leq 2 + 8\frac{(L+1) \sqrt{d} \sqrt{\log(K/\delta)}}{\Sigma} + 2LM d w_{k}^{\frac{1}{d+1}} \frac{(L+1) \sqrt{d} \sqrt{\log(K/\delta)}}{\Sigma^2} +d \\
&\leq 10 (L+1) \sqrt{d} \sqrt{\log(K/\delta)} (1 + \frac{1}{\Sigma^2}).
\end{align*}

By plugging in Equation~\ref{eq:alfin} the definition of $C$ and the bound on $B$ computed above, we obtain
\begin{align*}
 \sum_{i=1}^K \sum_{i=1}^{S_{k}} w_{k,i}^2 \si_{k,i}^2 
\leq&  \frac{1}{n^{\frac{d+2}{d}}} \Big[ \Sigma^{\frac{2(d+1)}{d}} +  650 M(L+1) d^{3/2} \Big(1 +  \frac{3M d}{\Sigma} \Big)^4  \sqrt{\log(K/\delta)} K^{\frac{1}{d+1}} n^{-\frac{1}{d+1}}\\
&+ 25 Md (\Sigma+1)^{\frac{2(d+1)}{d}}\Big(1 +  \frac{3M d}{\Sigma} \Big)^4 \big(\frac{1}{K}\big)^{\frac{1}{d+1}} \Big].
\end{align*}
This concludes the proof.

\end{document}